\documentclass[final]{cvpr}

\usepackage[OT1]{fontenc} 

\usepackage{times}
\usepackage{epsfig}
\usepackage{graphicx}
\usepackage{amsmath}
\usepackage{amssymb}
\usepackage{mathtools}
\usepackage{multirow}
\usepackage{booktabs}
\usepackage{array}
\usepackage{tabularx}
\usepackage[belowskip=-5pt]{caption}
\usepackage{subcaption}
\usepackage[table]{xcolor}
\definecolor{road}{rgb}{.502,.251,.502}
\definecolor{sidewalk}{rgb}{.957,.137,.910}
\definecolor{building}{rgb}{.275,.275,.275}
\definecolor{wall}{rgb}{.4,.4,.612}
\definecolor{fence}{rgb}{.745,.6,.6}
\definecolor{pole}{rgb}{.6,.6,.6}
\definecolor{tlight}{rgb}{.980,.667,.118}
\definecolor{tsign}{rgb}{.863,.863,0}
\definecolor{vegetation}{rgb}{.420,.557,.137}
\definecolor{terrain}{rgb}{.596,.984,.596}
\definecolor{sky}{rgb}{0,.510,.706}
\definecolor{person}{rgb}{.863,.078,.235}
\definecolor{rider}{rgb}{1,0,0}
\definecolor{car}{rgb}{0,0,.557}
\definecolor{truck}{rgb}{0,0,.275}
\definecolor{bus}{rgb}{0,.235,.392}
\definecolor{train}{rgb}{0,.314,.392}
\definecolor{motorbike}{rgb}{0,0,.902}
\definecolor{bicycle}{rgb}{.467,.043,.125}
\definecolor{unlabelled}{rgb}{0,0,0}

\usepackage[pagebackref=true,breaklinks=true,colorlinks,bookmarks=false]{hyperref}

\def\cvprPaperID{6} 
\def\confName{CVPR}
\def\confYear{2021}

\DeclareMathOperator{\ASR}{ASR}
\DeclareMathOperator{\mASR}{mASR}
\DeclareMathOperator{\IoU}{IoU}

\renewcommand\footnotemark{}

\begin{document}

\title{Latent Space Regularization for Unsupervised Domain Adaptation\\ in Semantic Segmentation}


\author{Francesco Barbato, Marco Toldo\thanks{Our work was in part supported by the Italian Ministry for Education (MIUR) under the ``Departments of Excellence" initiative (Law 232/2016) and by the SID project ``Semantic Segmentation in the Wild".}, Umberto Michieli, Pietro Zanuttigh\\
Department of Information Engineering, University of Padova\\
{\tt\small \{barbatofra, toldomarco, umberto.michieli, zanuttigh\}@dei.unipd.it}}

\maketitle

\begin{abstract}
Deep convolutional neural networks for semantic segmentation achieve outstanding accuracy, however they also have a couple of major drawbacks: first, they do not generalize well to  distributions slightly different from the one of the training data; second, they require a huge amount of labeled data for their optimization. In this paper, we introduce feature-level space-shaping regularization strategies to reduce the domain discrepancy in semantic segmentation. In particular, for this purpose we jointly enforce a clustering objective, a perpendicularity constraint and a norm alignment goal on the feature vectors corresponding to source and target samples. Additionally, we propose a novel measure able to capture the relative efficacy of an adaptation strategy compared to supervised training. We verify the effectiveness of such methods in the autonomous driving setting  achieving state-of-the-art results in multiple synthetic-to-real road scenes benchmarks.
\end{abstract}


%
%
%
\section{Introduction}
\label{subsec:intro}
Semantic segmentation is a key tool to address challenging scene understanding problems, like those connected to autonomous driving.
Reliable solutions have started to appear  with the rise of deep learning: most state-of-the-art approaches exploit Convolutional Neural Networks (CNNs) with an encoder-decoder architecture, starting from the pioneering FCN work~\cite{long2015} up to recent highly performing schemes like 
 PSPNet~\cite{zhao2017} and DeepLab~\cite{chen2017rethinking,chen2018encoder,chen2018deeplab}. 
These architectures typically exploit CNNs originally designed for image classification as feature extractors, 
removing their tailing fully-connected layers and adding a decoding network to generate full-size segmentation maps. 
The outstanding results of these models come at the cost of an expensive training process requiring
 massive amounts of labeled data. 
In spite of this, their generalization properties are not always satisfactory. When applied 
on domains similar but not identical to the training one, 
the models suffer from significant accuracy degradation: the so-called domain shift issue. 
%

To alleviate this problem, many Unsupervised Domain Adaptation (UDA) solutions have been proposed, exploiting unlabeled samples from the target domain to aid the generalization aptitude of the network.
The adaptation can be performed at different stages of the network, \ie, at the input, feature or output level~\cite{toldo2020unsupervised}. 
Deep  networks typically solve complex tasks by building some compact latent representations of the inputs, which are representative of the classifier output. These internal representations are extremely meaningful for the subsequent decision process~\cite{bengio2013representation,girshick2014rich}; nevertheless current UDA approaches for semantic segmentation hardly operate at this level due to the high dimensionality of the latent space.
We propose a novel strategy working at the less-explored feature level: our aim is to reduce the performance discrepancy by employing latent-space shaping objectives between source and target domains.

First of all, a clustering-based objective forces the feature vectors of each class to be closer to the corresponding prototype centroids. 
While based on source supervision for prototype estimation, its action is delivered to both source and target representations to achieve class-conditional domain alignment.
Then, an additional component enforces the perpendicularity of class prototypes, 
thus assembling features into well-distanced class clusters and, at the same time, promoting disjoint activation sets between semantic categories. 
Finally, we account for the fact that, as  noticed in \cite{Xu2019}, feature vectors computed from target domain samples tend to have smaller norms than source domain ones. 
This latter claim is due to domain-specific features, which the network relies on to solve the source-supervised classification. Yet, those features may be missing in the target domain and, therefore, may lead to a weakened response of neuron activations in target latent representations.
To address this issue, we introduce a regularization objective that promotes uniform vector norms across source and target representations, while jointly inducing progressively increased norm values.
Furthermore, the inter-class norm alignment has shown to remove distribution biases towards the most frequent classes, whose higher classification confidence is typically accompanied by bigger feature norms.

Since the proposed techniques require to set  a strong relationship between predicted segmentation maps and feature representations, we additionally develop a novel strategy to propagate semantic information from the labels to the lower 
resolution  feature space. 

\section{Related Works}
\label{subsec:related}

\textbf{Unsupervised Domain Adaptation} is a challenging setting in domain adaptation where only unlabeled samples from the target domain are used. 
The goal is to limit the performance degradation due to the distribution discrepancy between source and target data (\ie, the \textit{domain shift}~\cite{SunFS16}), with no supervision in the target domain.
%

Early techniques 
focused on whole-image classification 
\cite{borgwardt2006integrating,shotton2008semantic,vezhnevets2010,pathak2014fully}, while,
recently, the domain adaptation field has witnessed a rapid increase in interest, resulting in a multitude of different approaches for various tasks.
From a general point of view, we can identify three major categories of works~\cite{toldo2020unsupervised}, 
according to the network location in which they act, namely: input-level, feature-level and output-level.\\
Usually,  adaptation at the input level  aims to reduce the domain shift by acting directly on input images and trying to match source and target visual appearance (\textit{i.e.}, low level feature distribution), typically using generative adversarial schemes~\cite{chen2019crdoco,hoffman2018, hoffman2016,MurezKKRK18,toldo2020,pizzati2020domain}. 
%
At the output level, self-training concepts~\cite{zou2018,Zou2019} have been explored, where target network predictions in the form of pseudo-labels guide the learning process in a self-supervised manner. Alternatively, some works introduce entropy minimization techniques~\cite{Chen2019,vu2019advent}, which force the network to be more confident in the segmentation of target samples, thus mimicking the behavior shown in the source domain. Other approaches~\cite{biasetton2019,michieli2020adversarial,spadotto2020unsupervised} further exploit an adversarial discriminator to reduce the perceived discrepancy between segmentation maps produced by source and target domains. \\
A different line of works operates at the feature level, \textit{e.g.}, by enforcing the extraction of more discriminative features. 
For example, some works resort to dropout regularization~\cite{Lee2019,Park2018,Saito2018ADR} (either channel-wise or point-wise) to push decision boundaries away from high density regions, while others opt for domain adversarial feature alignment~\cite{du2019ssfdan, sankaranarayanan2018,tsai2018,Tsai2019} either by acting on them directly or by training the network on reconstructed images. 


\textbf{Latent Space Regularization} is a family of techniques that can be used to reduce the domain shift and have been applied in many semantic segmentation tasks such as UDA \cite{kang2019contrastive,tian2020domain}, continual learning \cite{michieli2021continual} and few-shot learning \cite{dong2018few,wang2019panet}.
In general, strategies belonging to this class make use of additional constraints imposed on the feature vectors, effectively reducing the extent of space each of them can occupy. Such reduction 
has shown to promote more overlap between the source and target distributions, thus reducing the domain shift \cite{toldo2020clustering}. 
In UDA, these techniques are generally applied in class-conditional manner, 
hence relying on the exclusive supervision of source samples.
However, it is reasonable to assume that their effect is reflected also on target samples.
These techniques have been exploited in various ways.
Different kinds of automatic feature clustering by embedding variations of the K-Means algorithm in the training procedure have been proposed \cite{kang2019contrastive,liang2019distant,wang2019unsupervised,tian2020domain}.
In \cite{toldo2020clustering} the authors further refine this idea by proposing an explicit clustering objective between feature vectors and the appropriate class prototypes. 
Another work~\cite{choi2020role} proposes feature-level orthogonality as an alternative for the standard cross-entropy optimization objective in image classification, trying to reduce the number of redundant features extracted by a CNN. Approaches closer to our strategy are \cite{pinheiro2018unsupervised}, where orthogonal class prototypes are used as a medium through which classification is performed in an unsupervised domain adaptation setting, and \cite{wu2019improving}, where an orthogonality constraint over the prototypes is exploited. 
\section{Problem Setup}
\label{subsec:setup}
In this section we 
detail our setup.
Formally, we denote the input image space as ${\mathcal{X} \! \subset \! \mathbb{R}^{H \times W \times 3}}$ and the associated output label space as ${\mathcal{Y} \! \subset \! \mathcal{C}^{H \times W}}$, where $H$ and $W$ represent the spatial dimensions and $\mathcal{C}$ the set of classes. 
Given a first training set ${ \mathcal{T}^s = \{ (\mathbf{X}_{n}^s,  \mathbf{Y}_{n}^s) \}_{n=1}^{N_s} }$, where labeled samples ${ (\mathbf{X}_{n}^s,  \mathbf{Y}_{n}^s) \in \mathcal{X}^s \times \mathcal{Y}^s }$ originate from a supervised source domain, together with a second set of unlabeled input samples ${ \mathcal{T}^t = \{ \mathbf{X}_{n}^t \}_{n=1}^{N_t} }$, from a target domain ($\mathbf{X}_{n}^t \in \mathcal{X}^t$), our goal is to transfer knowledge on the segmentation task learned on the source domain to the unsupervised target domain (\ie, without any label on the target set). Superscripts $s$ and $t$ specify the domain: source and target, respectively.

We assume that the segmentation network $S = D \circ E$ is based on an encoder-decoder architecture (as most recent approaches for semantic segmentation), \ie, made by the concatenation of two logical blocks: the encoder network $E$, consisting of the feature extractor, and a decoder network $D$, which is the actual classifier producing the segmentation map. 
%
Moreover, we call the features extracted from a generic input image $\mathbf{X}$ as $E(\mathbf{X}) = \mathbf{F} \in \mathbb{R}^{H' \times W' \times K}_{0+}$, where $K$ denotes the number of channels and $H' \times W'$ denotes the low-dimensional latent spatial resolution.
%
%
%
Given the structure of encoder-decoder convolutional segmentation networks, we can assume that each class is mapped to a reference representation in the latent space, that should be as invariant as possible to the domain shift. The techniques that will be introduced in Section \ref {subsec:method} try to enforce this by comparing the extracted features with some \textit{prototypes} for the various classes. In the rest of this section we show how to associate feature vectors to semantic classes and how to compute the prototypes.

\textbf{Histogram-Aware Downsampling.} 
Since the spatial information of an image is mostly preserved while its content travels through an encoder-decoder network, we can infer a strict relationship between any feature vector and the semantic labeling of the corresponding image region. 

Therefore, the first step of the extraction process is to identify a way to propagate the labeling information to 
latent representations 
(decimation),
preserving the semantic content of the image region (window) associated to each feature vector. 
Otherwise, the generation of erroneous associations would significantly impair the estimation objective. 
%
%
%
For this task, we design a non-linear pooling function: instead of computing a simple subsampling 
(\eg, nearest neighbor
), we compute a frequency histogram over the labels of all the pixels in each window. 
Such histograms are then used to select appropriate classification labels for the downsampled windows, producing feature-level label maps $\{\mathbf{I}^{s,t}_n\}^{N_{s,t}}_{n=1}$. 
Specifically, the choice is made by selecting the label corresponding to the frequency peak in each window, if such peak is distinctive enough, \ie, if any other peak is smaller than $T_h$ times the biggest one (
in a similar fashion to the orientation assignment step in the SIFT feature extractor \cite{790410}). Empirically, we set $T_h =0.5$.
A key feature of this technique is its ability to introduce void-class samples 
when a considered window cannot be assigned to a unique class, \ie, it contains mixed classification labels.
This procedure can be naturally extended to 
pseudo-labels (\ie, network-generated segmentation maps) 
via a confidence measure over the 
maps 
that preserves only reliable predictions.
In our case, such measure is computed efficiently by average pooling over the 
map of output probability peaks 
and used to mask the raw low-resolution pseudo-labels, \ie, we select only confident labeling with average probability value greater than 
$T_p = 0.5$, empirically.

\textbf{Prototype Extraction.} Once computed, the feature-level label maps ${\{\mathbf{I}^{s,t}_n\}^{N_{s,t}}_{n=1}}$ can be used to extract the set $\mathcal{F}_c$ of feature vectors belonging to a generic class $c \in \mathcal{C}$ in a training batch $\mathcal{B}$:
\begin{equation}
\label{eq:feats}
\mathcal{F}^{s,t}_c \! = \! \left\{ \mathbf{F}_n^{s,t}[h,w] \in \mathbb{R}^{K}_{0+} \mid \mathbf{I}^{s,t}_n[h,w] = c, \forall n \in \mathcal{B}\right\} ,
\end{equation}
where $[h,w]$ denote all possible spatial locations over a feature map, \ie, $0 \leq h < H'$  and $0 \leq w < W'$. 
Exploiting this definition, we can identify the set of all feature vectors in batch $\mathcal{B}$ as the union $\mathcal{F}^{s,t} = ( \bigcup_{c} \mathcal{F}^{s,t}_c ) \cup\mathcal{F}^{s,t}_v$ where $\mathcal{F}^{s,t}_v$ are the sets of void-class samples.
The class-wise sets are then used to estimate the per-batch class prototypes on labeled source data by simply computing their centroids:
\begin{equation}
\label{eq:proto}
\mathbf{p}_c[i] = \frac{1}{|\mathcal{F}^s_c|}\sum\limits_{\mathbf{f} \in \mathcal{F}^s_c}{\mathbf{f}[i]} \;\;\;\;\; \forall i,\;\; 1 \leq i \leq K .
\end{equation}
Finally, to reduce estimation noise and obtain more stable and reliable prototypes, we apply exponential smoothing:
\begin{equation}
\label{eq:smooth}
\hat{\mathbf{p}}_c = \eta \hat{\mathbf{p}}_c' + (1-\eta)\mathbf{p}_c .
\end{equation}
Where $\hat{\mathbf{p}}_c$ and $\hat{\mathbf{p}}_c'$ are the estimates of class $c$ prototype respectively at current and previous optimization steps. We initialized $\hat{\mathbf{p}}_c=\mathbf{0}$ and empirically set $\eta=0.8$.
This strategy allows us to keep track of classes that are not present in the current batch of source samples (in this case we set $\eta=1$ to propagate the previous estimate), aiding significantly in the unsupervised target tasks.
\section{Method}
\label{subsec:method}

\begin{figure*}[t]
\centering
\includegraphics[width=0.95\textwidth]{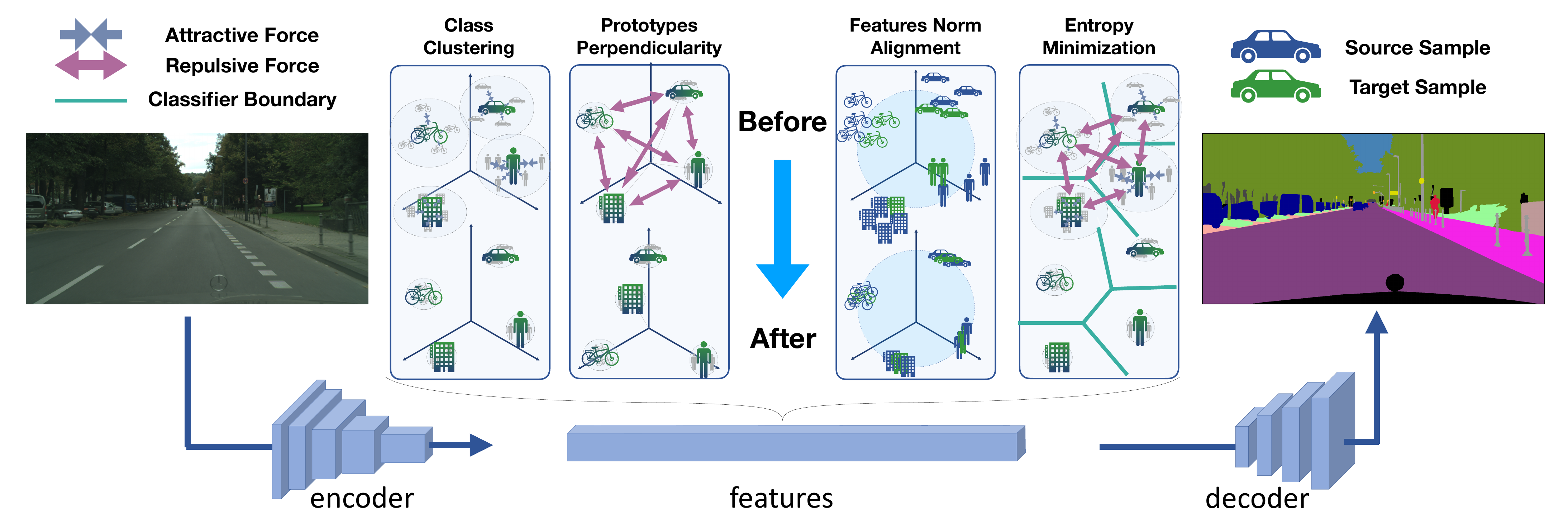}
\caption{Visual summary of our strategy and of the effect of its application on the feature space vectors. The three proposed space shaping constraint are from left to right: Class Clustering (\ref{loss:clust}),  Prototypes Perpendicularity (\ref{loss:perp}), Norm Alignment and Enhancement (\ref{loss:norm}). Furthermore we apply also entropy minimization \cite{Chen2019}.}
\label{fig:architecture}
\end{figure*}


In this section, we provide a detailed description of our approach, 
which is based on the idea of aiding the standard cross-entropy loss with additional components to enforce the regularization of the latent space. 
While the source supervised cross-entropy provides task discriminativeness to the model, the additional objectives jointly imposed on source and target representations drive towards feature-level domain invariance, ultimately reducing domain bias.
In particular, we add three feature-space shaping constraints
to the standard source-supervised cross-entropy loss $\mathcal{L}_{CE}^s$, whose combined effect can be mathematically expressed by:
\begin{equation}
\mathcal{L} = \mathcal{L}_{CE}^{s} + \lambda_C \cdot \mathcal{L}_C^{s,t} + \lambda_P \cdot \mathcal{L}_P^{s} + \lambda_N \cdot  \mathcal{L}_N^{s,t} .
\label{eq:tot}
\end{equation}
%
%
Here, $\mathcal{L}_C$ is the clustering loss on latent representations (Sec.~\ref{loss:clust}), $\mathcal{L}_P$ is the perpendicularity of class prototypes loss (Sec.~\ref{loss:perp}) and $\mathcal{L}_N$ is the norm alignment and enhancement loss (Sec.~\ref{loss:norm}). 
For ease of notation, Eq.\ \ref{eq:tot} reports each loss term once (the $s,t$ superscript here indicates the sum of source and target loss instances).
For an improved performance and to show that our approach can be applied on top of existing methods, we also extended our objective with the entropy minimization strategy proposed in \cite{Chen2019}, leading to $\mathcal{L}^+ = \mathcal{L} + \lambda_{EM} \cdot \mathcal{L}_{EM}$.\\ 
An overview of the complete approach is reported in Fig.~\ref{fig:architecture}.
\subsection{Clustering of Latent Representations}
\label{loss:clust}
The domain shift between source and target data is reflected into a discrepancy in distribution of latent representations from separate domains.
Moreover, as the lack of target supervision inherently leads to a bias towards the source domain, it is very likely for the classifier to trace decision boundaries tight around source embeddings, regardless of the disposition of unlabeled target instances.

Thus, the misalignment of class-conditional feature statistic inevitably leads the model towards incorrect classification over target representations, in turn degrading the segmentation accuracy on the target domain.
To cope with this issue, we start by introducing a clustering objective over the latent space, in order to achieve class-conditional alignment of feature distribution. 
By exploiting the prototypes introduced in Sec.~\ref{subsec:setup} 
and forcing source and target feature vectors to tightly assemble around 
them, we regularize the structure of the latent space, adapting representations into a domain independent 
class-wise distribution.

Mathematically, we define the clustering objective as:
\begin{equation}
 \mathcal{L}_C^{s,t} = 
 \frac{1}{|\mathcal{C}|} \sum\limits_{c \in \mathcal{C}} 
 \frac{1}{|\mathcal{F}^{s,t}_c|} 
 \sum\limits_{\mathbf{f} \in \mathcal{F}^{s,t}_c} 
 ||\hat{\mathbf{p}}_c-\mathbf{f}||^2,
\end{equation}%
where $||\cdot||$ denotes the L2 norm. 
%
%
%
%
This loss has multiple purposes: first, to better cluster representations in the latent space in a supervised manner, thus reducing the probability of erroneous classification. Second, to perform semi-supervised clustering on target samples exploiting network predictions as pseudo-labels
(notice that we used only the most confident labeling as detailed in Sec.\ \ref{subsec:setup}).
Finally, to improve prototype estimates, since forcing tighter clusters will result in more stable 
batch-wise centroids, which will be closer to the moving-averaged prototypes.

\subsection{Perpendicularity of Latent Representations}
\label{loss:perp}

We further enhance the space shaping action induced by the clustering objective by introducing a prototype perpendicularity loss.
The idea is to improve the segmentation accuracy by better separating the tight and domain-invariant clusters on both domains. 
By doing so, we allow the classifiers to increase the margin between decision boundaries and feature clusters, and, consequently, we reduce the likelihood of those boundaries to cross target high-density regions of the feature space (\ie, regions populated by many target samples). 
Unlike previous works \cite{toldo2020clustering}, class clusters of latent embeddings are forced to adopt a regular disposition, whilst being spaced out. 
In fact, we directly encourage a class-wise orthogonality property, by pushing prototypes to be perpendicular.
In this way, not only we increase the distance among class clusters, but we jointly regularize the latent space and encourage channel-wise disjoint activations between different semantic categories. 
%

To quantify perpendicularity in the loss value, we exploit the inner product in the euclidean space and its relationship with the angle $\theta$ between two vectors $\mathbf{j}$ and $\mathbf{k}$, \ie, $\mathbf{j}\cdot\mathbf{k} = ||\mathbf{j}||\;||\mathbf{k}||\cos\theta$.
Minimizing their normalized product is equivalent to maximizing the angle between them, since feature vectors have non-negative values. 
To capture this, we enforce cross-perpendicularity between any couple of prototypes:
\begin{equation}
\mathcal{L}_P^{s} = \frac{1}{|\mathcal{C}|(|\mathcal{C}|-1)}\sum\limits_{c_i,c_j \in \mathcal{C}, i \neq j}{\frac{\mathbf{p}_{c_i}}{||\mathbf{p}_{c_i}||}\cdot\frac{\mathbf{p}_{c_j}}{||\mathbf{p}_{c_j}||}}.
\label{eq:perp}
\end{equation}
Eq.~\ref{eq:perp} computes the sum of the cosines over the set of all couples 
of non-void classes.
We use the per-batch computation of prototypes on source samples $\mathbf{p}_c$ (notice the missing hat on the prototype symbols, see Eq.~\ref{eq:proto}), 
guaranteeing a stronger gradient flow through the network.
In addition, thanks to the tight geometric relation between prototype estimates and feature vectors enforced by $\mathcal{L}_C^{s,t}$, the effect induced by the orthogonality constraint on the prototypes is propagated to the vectors associated to them.\\
The net result is the application of the shaping action to all feature vectors of each class, thus promoting perpendicularity between all individual components of distinct clusters.\\
The loss seeks to increase the angular distance between latent representations of separate classes, which is achieved when distinct sets of active feature channels are associated to distinct semantic categories.
A similar orthogonality constraint has been proposed in \cite{toldo2020clustering},
but here we directly enforce the perpendicularity property between clusters, whereas in \cite{toldo2020clustering} each feature vector is considered independently without accounting for its semantic labeling, as the constraint is imposed in an unsupervised fashion.

\subsection{Latent Norm Alignment Constraint}
\label{loss:norm}

The last constraint we propose acts on the 
norm of source and target feature vectors.
In particular, we promote the extraction of latent representations with uniform norm values across domains.
Our objective is twofold.
First, we aim at increasing the classification confidence during target prediction, similarly to what achieved by adaptation strategies based on entropy minimization over the output space \cite{vu2019advent}.
In particular, recent studies in image classification \cite{Xu2019} highlight how the norm of target feature vectors tends towards smaller values than source ones, generally leading to reduced prediction confidence and potentially erroneous classifications. 
Second, we assist the perpendicularity loss by reducing the number of domain-specific feature channels 
exploited to perform classification. 
We argue, in fact, that by forcing the network to produce consistent feature norms, we reduce the number of channel activations switched on for only one of the two domains, as they would cause norm discrepancies.
Formally, we define two separate objectives for source and target domains:
\begin{align}
\label{loss:normS}
&\mathcal{L}_N^{s} \! =\! \frac{1}{|\mathcal{F}^s|} \sum\limits_{\mathbf{f} \in \mathcal{F}^{s} } {\left| (\bar{f}_s+\Delta_f) \! - \! ||\mathbf{f}|| \right|}, \\
\label{loss:normT}
&\mathcal{L}_N^{t} \! =\! \frac{1}{|\mathcal{F}^{t}|} \sum\limits_{\mathbf{f} \in \mathcal{F}^{t}} {\max(0,{(\bar{f}_s+\Delta_f) \! - \! ||\mathbf{f}||})  },
\end{align}
where $\bar{f}_s$ is the mean of the feature vector norms computed from source samples in the previous optimization step and $\Delta_f$ dictates the enhancement step (we experimentally tuned it, \eg, $\Delta_f = 0.002$). 
Feature vectors are pushed towards the same global average norm value, regardless of their labeling. This removes any bias generated by heterogeneous pixel-class distribution in semantic labels, which, for example, would cause the most frequent classes to show larger norm than the average. 
The source-specific constraint (Eq.~\ref{loss:normS}) forces both the inter-class alignment and enhancement step, 
\ie, it ensures that 
norms are progressively increased throughout the training process, towards a common value for all the classes.
On the other hand, the target objective (Eq.~\ref{loss:normT}) focuses on domain-alignment, by enforcing the target norms to be similar to the source ones.
Furthermore, since target features have typically smaller norms, 
we do not penalize target norms exceeding the 
 reference value. 

\section{Implementation Details}
\label{subsec:implementation}
\textbf{Datasets and Setup.} We test our model (LSR, Latent Space Regularization) on synthetic-to-real UDA on road-view semantic segmentation.
As (synthetic) source domains we employ the \textit{GTAV}~\cite{Richter2016} and the \textit{SYNTHIA}~\cite{ros2016} datasets.
The former is comprised of $24,966$ densely labeled, high-resolution ($1914\times1052$ px) images, taken from a video sequence produced by the GTAV game engine, while the latter provides $9,500$ densely labeled samples with resolution $1280\times760$ px, produced using the homonym software.
As target domain, we choose the \textit{Cityscapes}~\cite{Cordts2016} dataset, which contains $5,000$ densely labeled high-resolution ($2048\times1024$ px) images, acquired in European cities.

We train the architecture in a closed-set~\cite{toldo2020unsupervised} setup, \ie, source and target class sets coincide.
Therefore, we use the $19$ and $16$ common classes for \textit{GTAV} and \textit{SYNTHIA}, respectively.
\textit{GTAV} and \textit{Cityscapes} images are resized during training to $1280\times720$ px  and $1024\times512$ px, respectively, while \textit{SYNTHIA} samples are kept at the original resolution. 

\textbf{Baseline Model.} We used the common ~\cite{tsai2018, vu2019advent, Chen2019, tranheden2020dacs, toldo2020clustering} DeepLabV2 network \cite{chen2017rethinking,chen2018encoder,chen2018deeplab}, with ResNet101~\cite{he2016deep} as the backbone (with $K=2048$ channels at the last level of the encoder) and stride $8$. 
We pre-train the model on source-only samples with a batch size of $10$ for more stable training, using patches of $512\times512$ px and data augmentation to remove visual biases introduced by the running mean components of batch-normalization layers when full images are employed. As an example, a dark patch on the bottom half of the image will often be interpreted as \textit{road}, while a light patch on the top half will often be interpreted as \textit{sky}, which is not always true (
see the random camera angles 
in the \textit{SYNTHIA} dataset) and preserving such behavior may be detrimental for some applications.
The final goal is to reduce color 
and texture-based biases that could be introduced during training on a single source dataset.

\textbf{Training Procedure.} We optimize the network using SGD with 
momentum of rate $0.9$ and weight decay regularization of $5 \times 10^{-4}$. The learning rate was scheduled according to a polynomial decay of power $0.9$ starting from $2.5 \times 10^{-4}$ over $250k$ steps, following~\cite{Chen2019}.
A subset of the original training set was  exploited as validation set for the hyper-parameters search in our loss terms.
To reduce overfitting we employ various dataset augmentation strategies: random left-right flip; white point re-balancing $\propto \mathcal{U}([-75,75])$
; color jittering $\propto \mathcal{U}([-25,25])$ (both applied independently over color channels) and random Gaussian blur~\cite{zou2018,Chen2019}.  
We used one NVIDIA Titan RTX GPU, with batch size of $2$ ($1$ source and $1$ target samples), training the network for $27,450$ steps (\ie, $10$ epochs of the Cityscapes~\cite{Cordts2016} dataset) and employing early stopping based on the validation set.


\newcommand{\Cdash}{\hspace*{.8em}\hbox{-}}
\newcommand{\lenA}{.45em}
\newcommand{\lenB}{.8em}
\newcommand{\lenC}{1.2em}
\newcolumntype{P}[1]{>{\fontsize{8}{8}\selectfont\centering\arraybackslash}p{#1}}
\newcolumntype{Q}[1]{>{\fontsize{8}{8}\selectfont}p{#1}}
\newcolumntype{Y}[1]{>{\fontsize{8}{8}\selectfont\centering\arraybackslash}X{#1}}
\begin{table*}[h!]
\linespread{0.7}\selectfont\centering
\begin{tabularx}{\textwidth}{P{\lenA}YQ{\lenA}Q{\lenA}Q{\lenA}Q{\lenA}Q{\lenA}Q{\lenA}Q{\lenA}Q{\lenA}Q{\lenA}Q{\lenA}Q{\lenA}Q{\lenA}Q{\lenA}Q{\lenA}Q{\lenA}Q{\lenA}Q{\lenA}Q{\lenA}Q{\lenB}P{\lenC}P{\lenC}P{\lenC}P{\lenC}}
\toprule[1pt]
\rotatebox{45}{Setup} & Configuration & \rotatebox{45}{Road} & \rotatebox{45}{Sidewalk} & \rotatebox{45}{Building} & \rotatebox{45}{Wall\textsuperscript{1}} & \rotatebox{45}{Fence\textsuperscript{1}} & \rotatebox{45}{Pole\textsuperscript{1}} & \rotatebox{45}{Traffic Light} & \rotatebox{45}{Traffic Sign} & \rotatebox{45}{Vegetation} & \rotatebox{45}{Terrain} & \rotatebox{45}{Sky} & \rotatebox{45}{Person} & \rotatebox{45}{Rider} & \rotatebox{45}{Car} & \rotatebox{45}{Truck} & \rotatebox{45}{Bus} & \rotatebox{45}{Train} & \rotatebox{45}{Motorbike} & \rotatebox{45}{Bicycle} & \makebox{mIoU} & \makebox{mIoU\textsuperscript{1}} & \makebox{mASR} & \makebox{mASR\textsuperscript{1}}\\
\noalign{\smallskip} 
\toprule[1pt]
& Target Only & 96.5 & 73.8 & 88.4 & 42.2 & 43.7 & 40.7 & 46.1 & 58.6 & 88.5 & 54.9 & 91.9 & 68.7 & 46.2 & 90.7 & 68.8 & 69.9 & 48.8 & 47.6 & 64.5 & 64.8 & - & 100 & 100\\
\toprule[1pt]
\footnotesize
\multirow{8}{*}{\rotatebox{90}{\quad \ \  From GTAV}} & Baseline~\cite{toldo2020clustering} & 71.4 & 15.3 & 74.0 & 21.1 & 14.4 & 22.8 & 33.9 & 18.6 & 80.7 & 20.9 & 68.5 & 56.6 & 27.1 & 67.4 & 32.8 & 5.6 & 7.7 & 28.4 & 33.8 & 36.9 & - & 54.0 & -\\
\cline{2-25}
\noalign{\smallskip}
& ASN (feat)~\cite{tsai2018} & 83.7 & 27.6 & 75.5 & 20.3 & 19.9 & 27.4 & 28.3 & 27.4 & 79.0 & 28.4 & 70.1 & 55.1 & 20.2 & 72.9 & 22.5 & 35.7 & \textbf{8.3} & 20.6 & 23.0 & 39.0 & - & 56.9 & -\\
& MinEnt~\cite{vu2019advent} & 84.4 & 18.7 & 80.6 & 23.8 & 23.2 & 28.4 & \underline{36.9} & 23.4 & 83.2 & 25.2 & \underline{79.4} & 59.0 & \textbf{29.9} & 78.5 & 33.7 & 29.6 & 1.7 & 29.9 & 33.6 & 42.3 & - & 61.9 & -\\
& SAPNet~\cite{li2020spatial} & \underline{88.4} & \textbf{38.7} & 79.5 & \underline{29.4} & \textbf{24.7} & 27.3 & 32.6 & 20.4 & 82.2 & 32.9 & 73.3 & 55.5 & 26.9 & \underline{82.4} & 31.8 & 41.8 & 2.4 & 26.5 & 24.1 & 43.2 & - & 63.1 & -\\
& MaxSquareIW~\cite{Chen2019} & 87.7 & 25.2 & \textbf{82.9} & \textbf{30.9} & \underline{24.0} & \underline{29.0} & 35.4 & 24.2 & \underline{84.2} & \underline{38.2} & 79.2 & 59.0 & \underline{27.7} & 79.5 & 34.6 & 44.2 & \underline{7.5} & \textbf{31.1} & \textbf{40.3} & 45.5 & - & 62.2 & -\\
& UDA OCE~\cite{toldo2020clustering} & \textbf{89.4} & 30.7 & 82.1 & 23.0 & 22.0 & \textbf{29.2} & \textbf{37.6} & \textbf{31.7} & 83.9 & 37.9 & 78.3 & \textbf{60.7} & 27.4 & \textbf{84.6} & \underline{37.6} & \underline{44.7} & 7.3 & 26.0 & 38.9 & \underline{45.9} & - & \underline{67.3} & -\\
& LSR (Ours) & 87.7 & \underline{32.6} & \underline{82.6} & 29.1 & 23.0 & 28.5 & 36.1 & \underline{28.5} & \textbf{84.8} & \textbf{41.8} & \textbf{80.1} & \underline{59.4} & 23.8 & 76.5 & \textbf{38.4} & \textbf{45.8} & 7.1 & 28.5 & \underline{40.1} & \textbf{46.0} & - & \textbf{67.7} & -\\
\toprule[1pt]
\multirow{8}{*}{\rotatebox{90}{\ \ From SYNTHIA}} & Baseline~\cite{toldo2020clustering} & 17.7 & 15.0 & 74.3 & 10.1 & 0.1 & 25.5 & 6.3 & 10.2 & 75.5 & \Cdash & 77.9 & 57.1 & 19.2 & 31.2 & \Cdash & 31.2 & \Cdash & 10.0 & 20.1 & 30.1 & 34.3 & 41.7 & 44.6\\
\cline{2-25}
\noalign{\smallskip}
& ASN (feat)~\cite{tsai2018} & 62.4 & 21.9 & 76.3 & \Cdash & \Cdash & \Cdash & 11.7 & 11.4 & 75.3 & \Cdash & 80.9 & 53.7 & 18.5 & 59.7 & \Cdash & 13.7 & \Cdash & \textbf{20.6} & 24.0 & - & 40.8 & - & 52.5\\
& MinEnt~\cite{vu2019advent} & 73.5 & 29.2 & 77.1 & 7.7 & \underline{0.2} & 27.0 & 7.1 & 11.4 & 76.7 & \Cdash & \underline{82.1} & \underline{57.2} & \underline{21.3} & 69.4 & \Cdash & 29.2 & \Cdash & 12.9 & 27.9 & 38.1 & 44.2 & 51.1 & 56.3\\
& SAPNet~\cite{li2020spatial} & \underline{81.7} & 33.5 & 75.9 & \Cdash & \Cdash & \Cdash & 7.0 & 6.3 & 74.8 & \Cdash & 78.9 & 52.1 & \underline{21.3} & 75.7 & \Cdash & 30.6 & \Cdash & 10.8 & 28.0 & - & 44.3 & - & 56.0\\
& MaxSquareIW~\cite{Chen2019} & 78.9 & 33.5 & 75.3 & \textbf{15.0} & \textbf{0.3} & \underline{27.5} & \underline{13.1} & \textbf{16.7} & 73.8 & \Cdash & 77.7 & 50.4 & 19.9 & 66.7 & \Cdash & \textbf{36.1} & \Cdash & 13.7 & \underline{32.1} & 39.4 & 45.2 & 53.8 & 58.3\\
& UDA OCE~\cite{toldo2020clustering} & \textbf{88.3} & \textbf{42.2} & \underline{79.1} & 7.1 & \underline{0.2} & 24.4 & \textbf{16.8} & \underline{16.5} & \textbf{80.0} & \Cdash & \textbf{84.3} & 56.2 & 15.0 & \textbf{83.5} & \Cdash & 27.2 & \Cdash & 6.3 & 30.7 & \underline{41.1} & \textbf{48.2} & \underline{54.3} & \underline{60.9}\\
& LSR (Ours) & 81.0 & \underline{36.9} & \textbf{79.5} & \underline{13.4} & \underline{0.2} & \textbf{28.7} & 9.0 & 16.1 & \underline{79.1} & \Cdash & 81.7 & \textbf{57.9} & \textbf{21.6} & \underline{77.2} & \Cdash & \underline{35.3} & \Cdash & \underline{14.2} & \textbf{35.4} & \textbf{41.7} & \underline{48.1} & \textbf{56.5} & \textbf{61.6}\\
\end{tabularx}
\caption{Comparison of adaptation strategies in terms of IoU, mIou and mASR \% (Sec.~\ref{subsec:results}). Best in \textbf{bold}, runner-up \underline{underlined}. mIoU\textsuperscript{1} and mASR\textsuperscript{1} restrict to 13 classes, ignoring the ones with same superscript. We use same setup and codebase of~\cite{Chen2019,toldo2020clustering}.}
\label{table:results}
\end{table*}

\section{Results}
\label{subsec:results}
%
%
In this section, we report the quantitative and qualitative results produced by the proposed approach (LSR), comparing it with several feature-level approaches~\cite{tsai2018,li2020spatial,toldo2020clustering} 
and with some works, like entropy minimization strategies~\cite{vu2019advent,Chen2019}, whose effect on feature distribution is found to be similar to ours. An ablation study and a discussion of the effects brought by the proposed loss terms is also included.

Notice that  our method is trained end-to-end, 
thus we can seamlessly add other adaptation techniques,
\eg, entropy minimization or adversarial input or output level approaches.
To prove such compatibility, 
 we introduce an entropy-minimization loss~\cite{Chen2019} in our framework.

%

\subsection{Measuring the Adaptation Performances}

For a better evaluation, we introduce a novel measure, called mASR (\textit{mean Adapted-to-Supervised Ratio}), to capture the relative performance between an adapted architecture and its target supervised counterpart.
 We start by computing the ratio between the IoU score of the adapted network for each class $c\in \mathcal{C}$ ($\IoU^c_{adapt}$) and the IoU of supervised training on target data ($\IoU^c_{sup}$), that is a reasonable upper bound estimate. Notice that higher means better, with a value of $1$ meaning that the adapted architecture has the same performances of supervised training. Finally we average the scores over all the classes in the  dataset:
\begin{equation}
\mASR = \frac{1}{|\mathcal{C}|} \sum\limits_{c \in \mathcal{C}}{\ASR^c}, \;\;\; \ASR^c \overset{def}{=} \frac{\IoU^c_{adapt}}{\IoU^c_{sup}} .
\end{equation}
Differently from other recent measures \cite{rezatofighi2019generalized,luiten2021hota}, the $\mASR$ measure captures the relative performance between an adapted architecture and its supervised counterpart, allowing an evaluation of  adaptation schemes more independent from the overall performances of the network on the various datasets. The per-class $\ASR$ values allows to quickly discover the most challenging classes for the adaptation task. Notice how, if compared with the standard mIoU, a key difference is that the contribution of each class is inversely proportional to the capacity of the segmentation model to learn it in the supervised reference scenario, thus emphasizing the most challenging semantic categories.
\newcommand{\imgWidth}{0.162\textwidth}
\begin{figure*}
\newcolumntype{Y}{>{\centering\arraybackslash}X}
\centering
\begin{subfigure}{.6em}
\scriptsize\rotatebox{90}{~~~GTAV$\rightarrow$Cityscapes}
\end{subfigure}%
\begin{subfigure}{\textwidth}
\vspace*{-.3em}
\centering
\begin{subfigure}{\imgWidth}
\centering
\includegraphics[width=\textwidth]{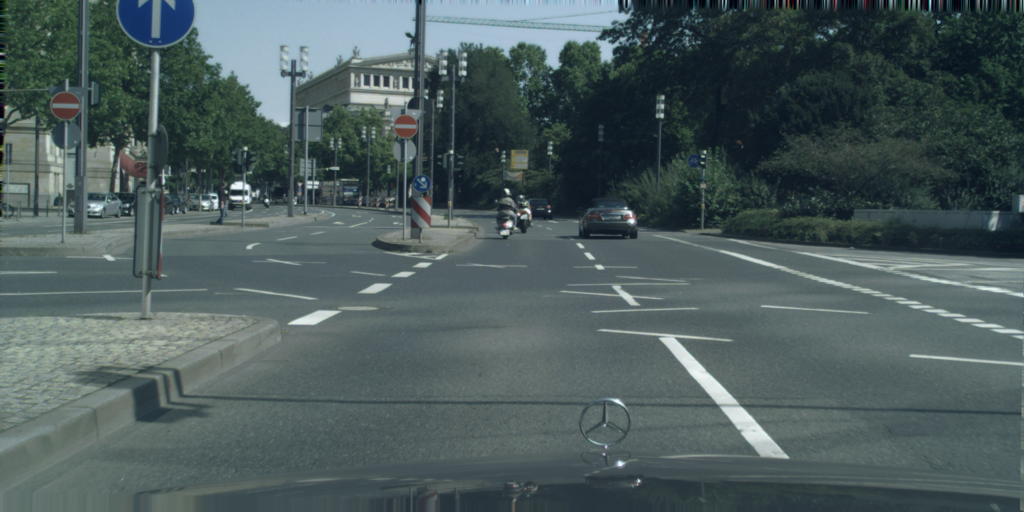}
\end{subfigure}%
\begin{subfigure}{\imgWidth}
\centering
\includegraphics[width=\textwidth]{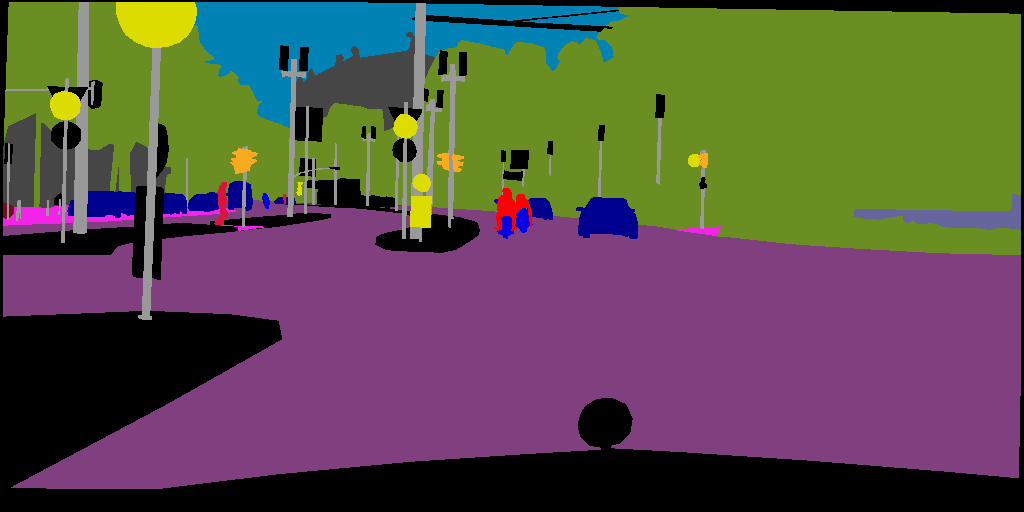}
\end{subfigure}%
\begin{subfigure}{\imgWidth}
\centering
\includegraphics[width=\textwidth]{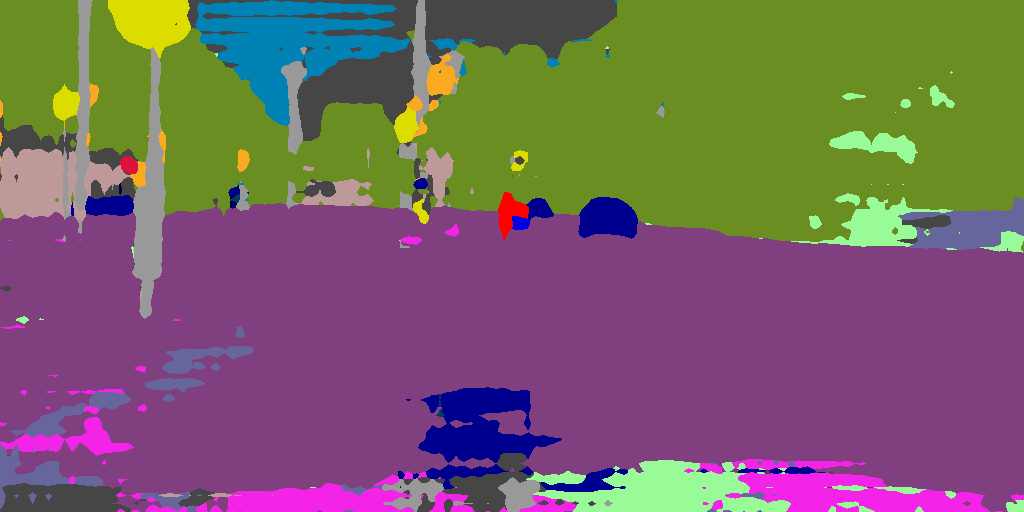}
\end{subfigure}%
\begin{subfigure}{\imgWidth}
\centering
\includegraphics[width=\textwidth]{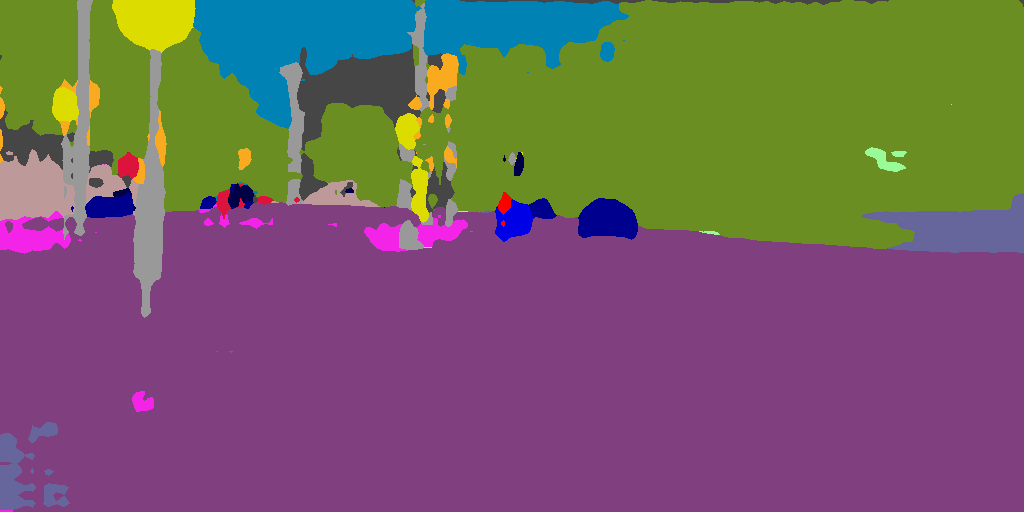}
\end{subfigure}%
\begin{subfigure}{\imgWidth}
\centering
\includegraphics[width=\textwidth]{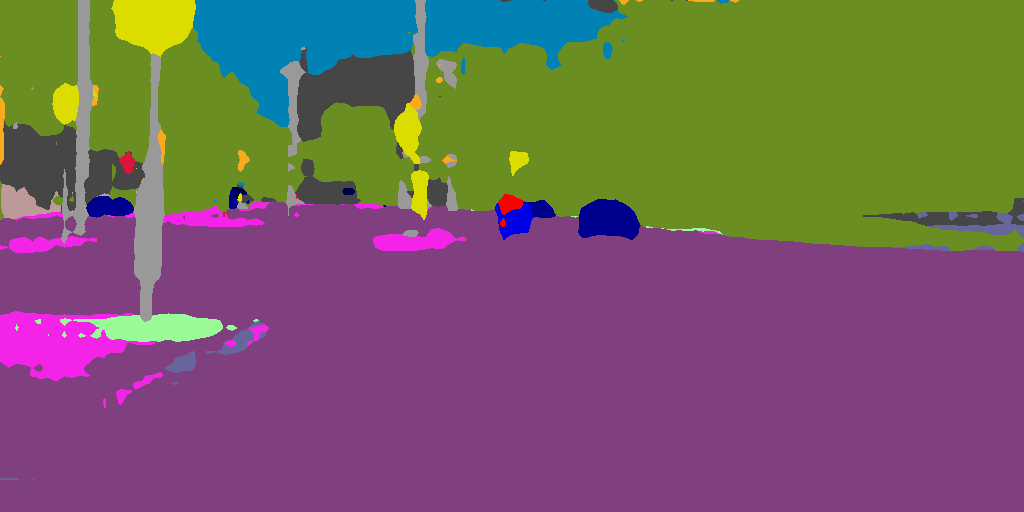}
\end{subfigure}%
\begin{subfigure}{\imgWidth}
\centering
\includegraphics[width=\textwidth]{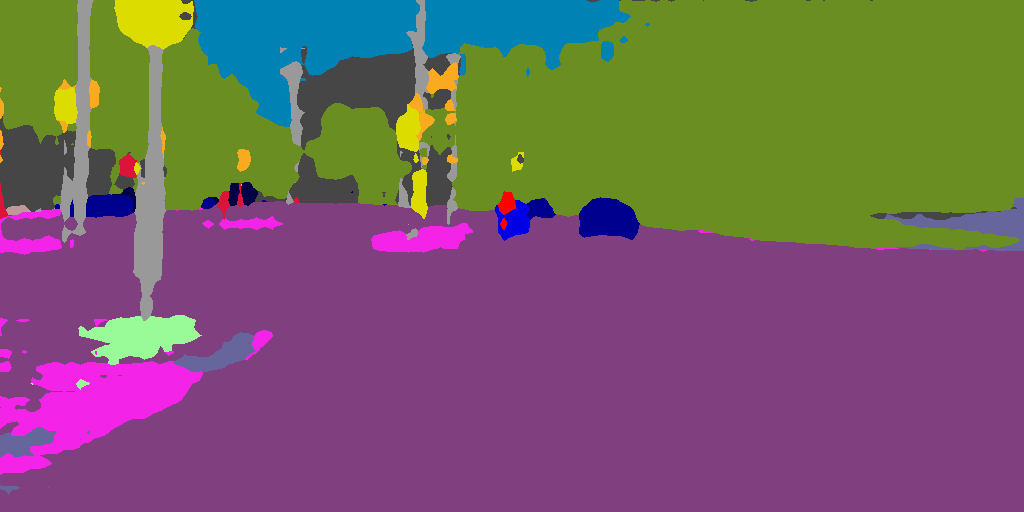}
\end{subfigure}

\begin{subfigure}{\imgWidth}
\centering
\includegraphics[width=\textwidth]{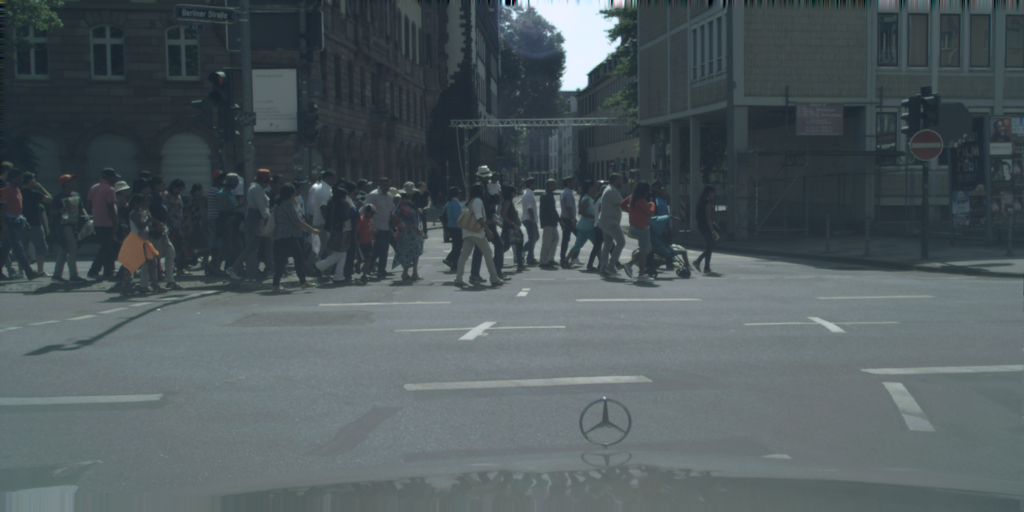}
\end{subfigure}%
\begin{subfigure}{\imgWidth}
\centering
\includegraphics[width=\textwidth]{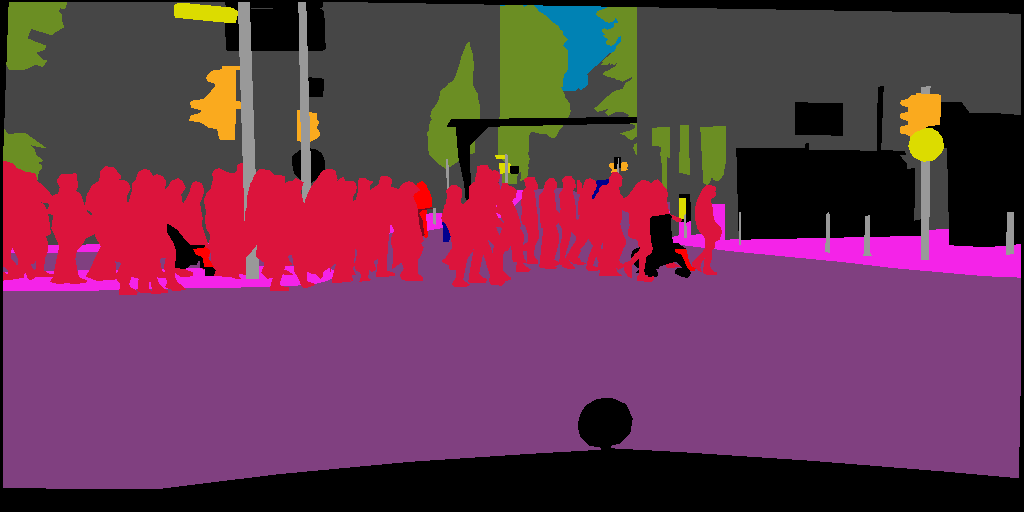}
\end{subfigure}%
\begin{subfigure}{\imgWidth}
\centering
\includegraphics[width=\textwidth]{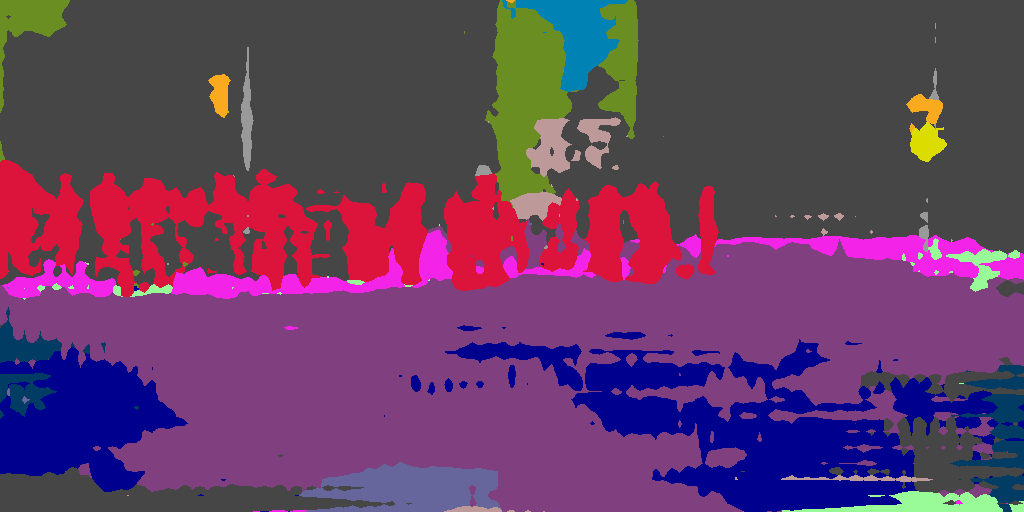}
\end{subfigure}%
\begin{subfigure}{\imgWidth}
\centering
\includegraphics[width=\textwidth]{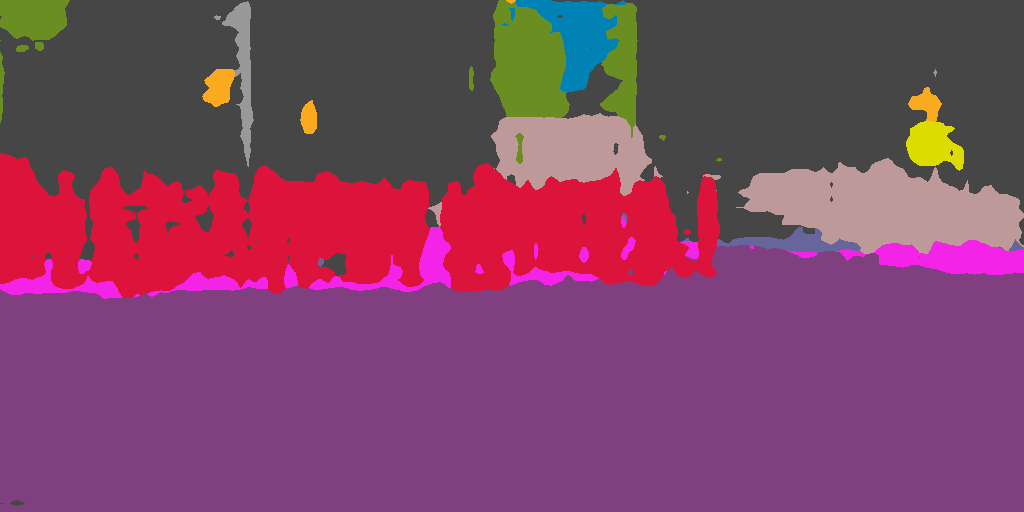}
\end{subfigure}%
\begin{subfigure}{\imgWidth}
\centering
\includegraphics[width=\textwidth]{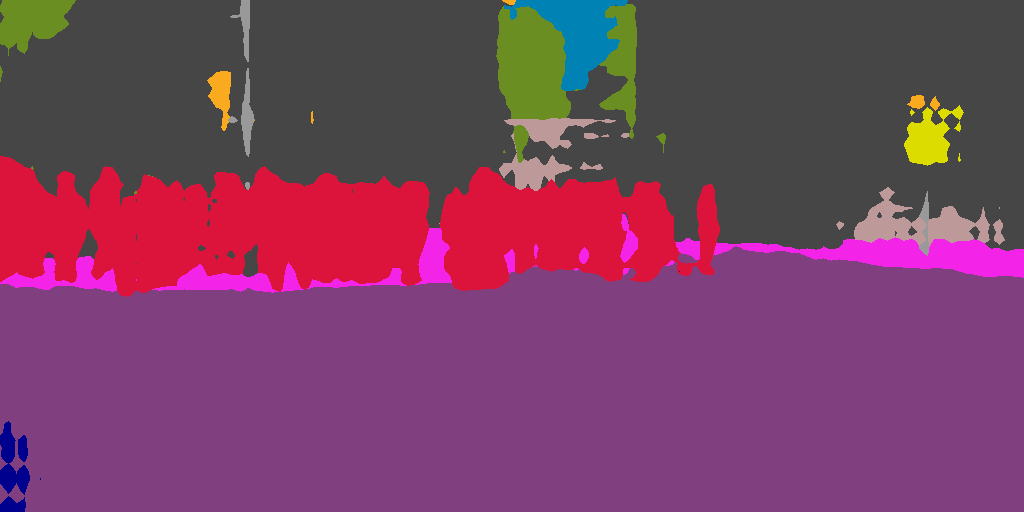}
\end{subfigure}%
\begin{subfigure}{\imgWidth}
\centering
\includegraphics[width=\textwidth]{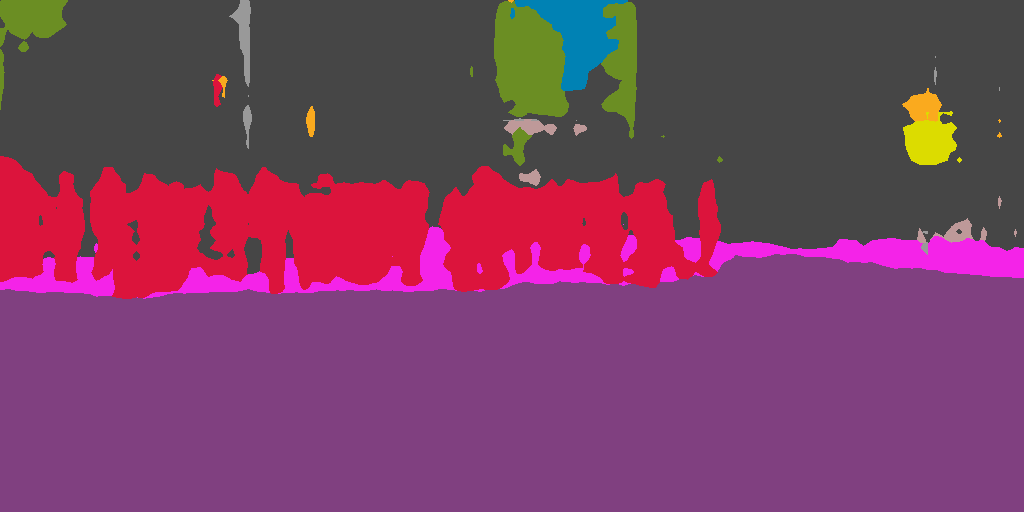}
\end{subfigure}%
\end{subfigure}
\begin{subfigure}{\textwidth}
\small
\begin{tabularx}{\textwidth}{YYYYYYYYYY}
\cellcolor{road} \textcolor{white}{Road} & \cellcolor{sidewalk} Sidewalk & \cellcolor{building} \textcolor{white}{Building} & \cellcolor{wall} \textcolor{white}{Wall} & \cellcolor{fence} Fence & \cellcolor{pole} Pole & \cellcolor{tlight} T. Light & \cellcolor{tsign} T. Sign & \cellcolor{vegetation} \textcolor{white}{\footnotesize Vegetation} & \cellcolor{terrain} Terrain \\
\cellcolor{sky} Sky & \cellcolor{person} \textcolor{white}{Person} & \cellcolor{rider} \textcolor{white}{Rider} & \cellcolor{car} \textcolor{white}{Car} & \cellcolor{truck} \textcolor{white}{Truck} & \cellcolor{bus} \textcolor{white}{Bus} &  \cellcolor{train} \textcolor{white}{Train} & \cellcolor{motorbike} \footnotesize \textcolor{white}{Motorbike} & \cellcolor{bicycle} \textcolor{white}{Bicycle} & \cellcolor{unlabelled} \footnotesize \textcolor{white}{Unlabeled}
\end{tabularx}
\end{subfigure}
\begin{subfigure}{.6em}
\scriptsize\rotatebox{90}{~~~~~SYNTHIA$\rightarrow$Cityscapes}
\end{subfigure}%
\begin{subfigure}{\textwidth}
\centering
\begin{subfigure}{\imgWidth}
\centering
\includegraphics[width=\textwidth]{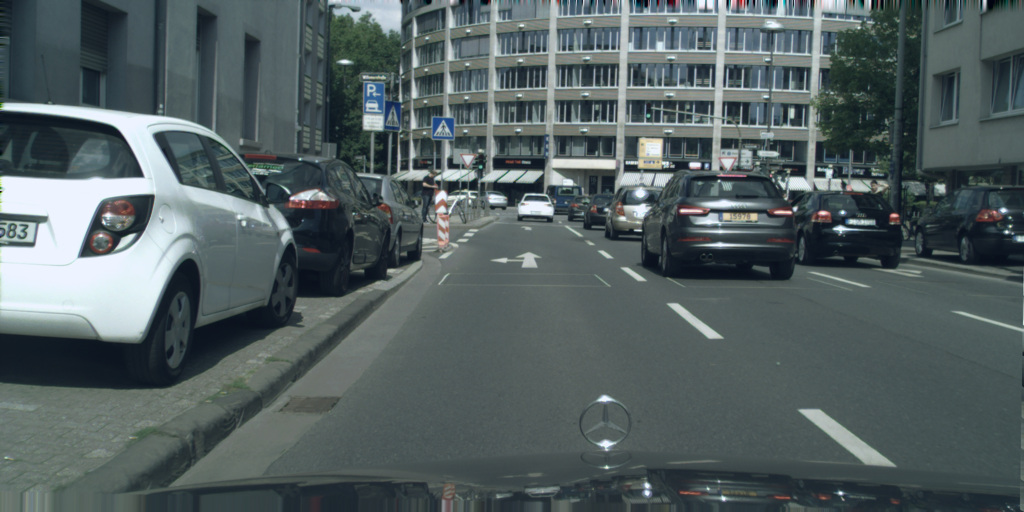}
\end{subfigure}%
\begin{subfigure}{\imgWidth}
\centering
\includegraphics[width=\textwidth]{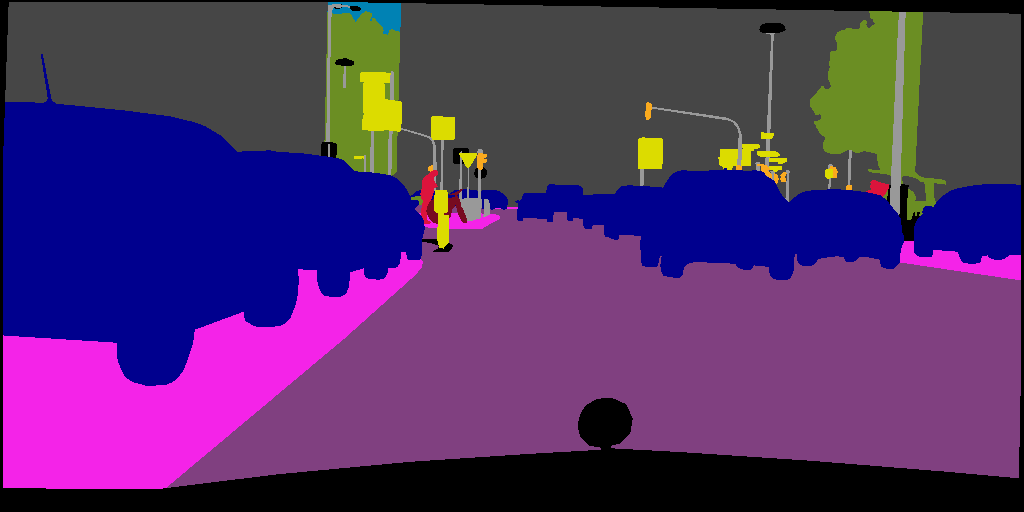}
\end{subfigure}%
\begin{subfigure}{\imgWidth}
\centering
\includegraphics[width=\textwidth]{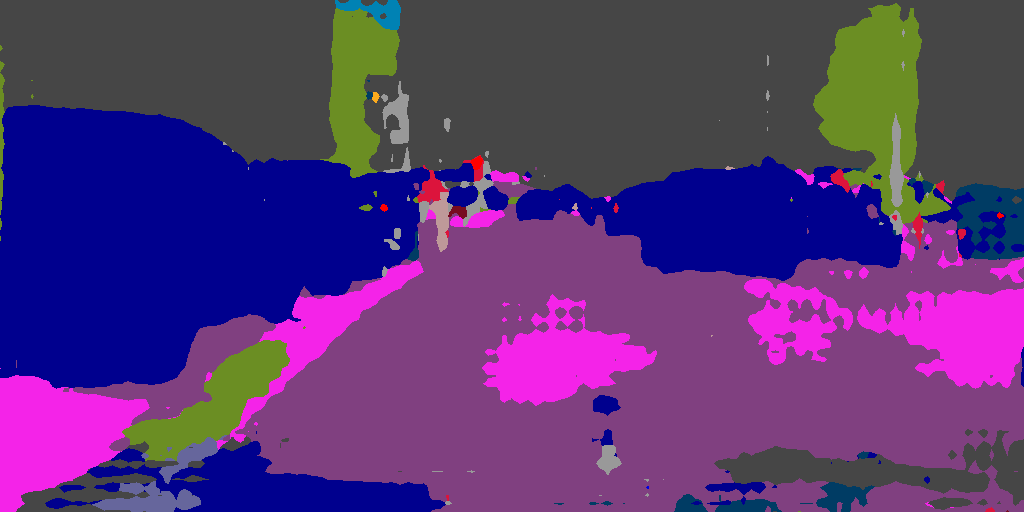}
\end{subfigure}%
\begin{subfigure}{\imgWidth}
\centering
\includegraphics[width=\textwidth]{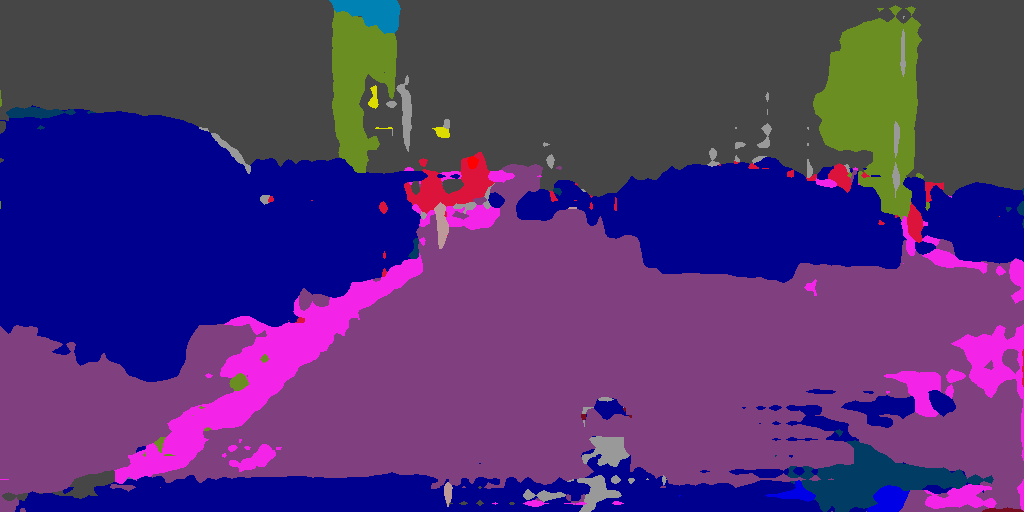}
\end{subfigure}%
\begin{subfigure}{\imgWidth}
\centering
\includegraphics[width=\textwidth]{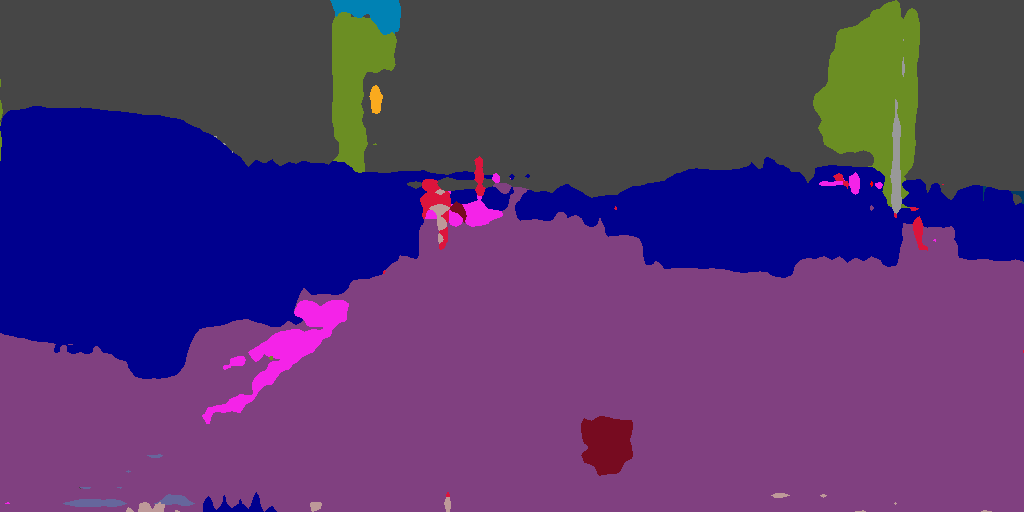}
\end{subfigure}%
\begin{subfigure}{\imgWidth}
\centering
\includegraphics[width=\textwidth]{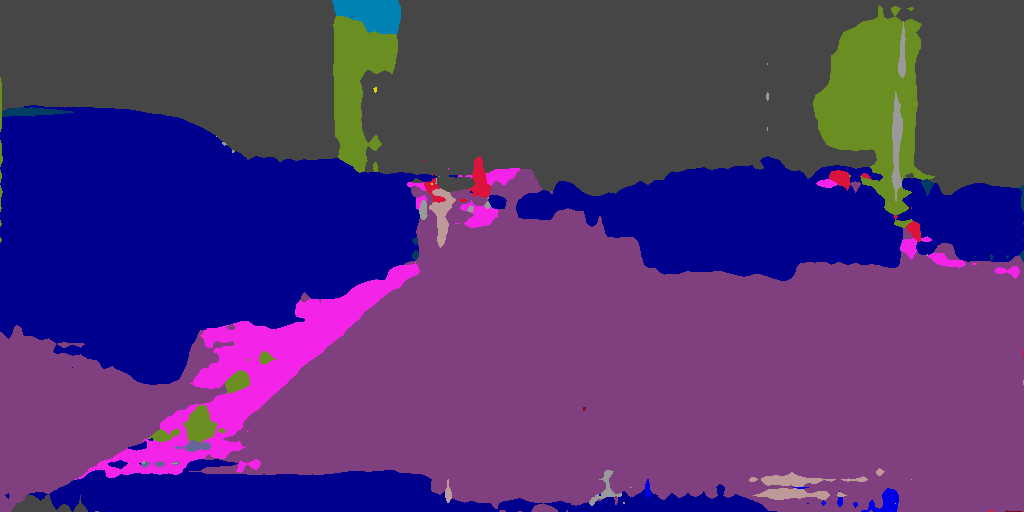}
\end{subfigure}

\begin{subfigure}{\imgWidth}
\centering
\includegraphics[width=\textwidth]{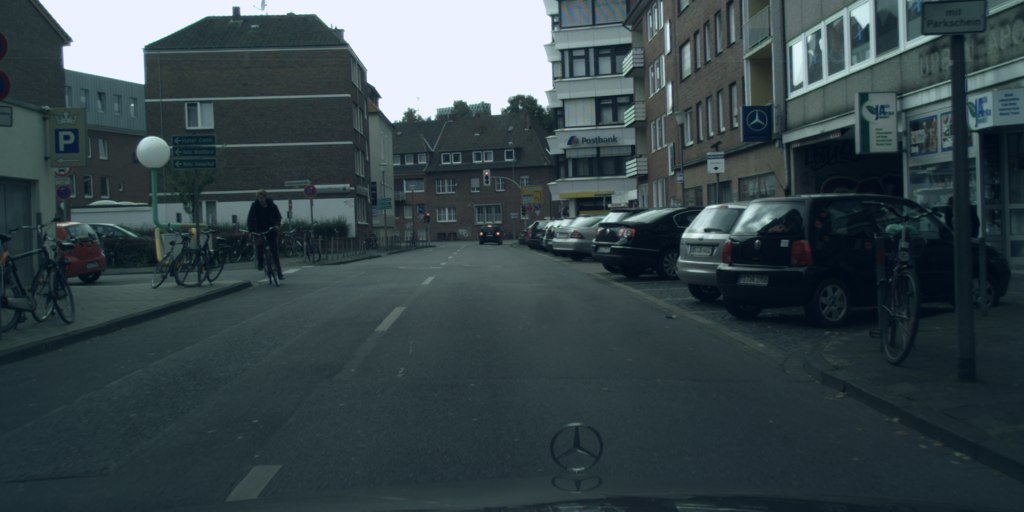}
\caption*{RGB Input}
\end{subfigure}%
\begin{subfigure}{\imgWidth}
\centering
\includegraphics[width=\textwidth]{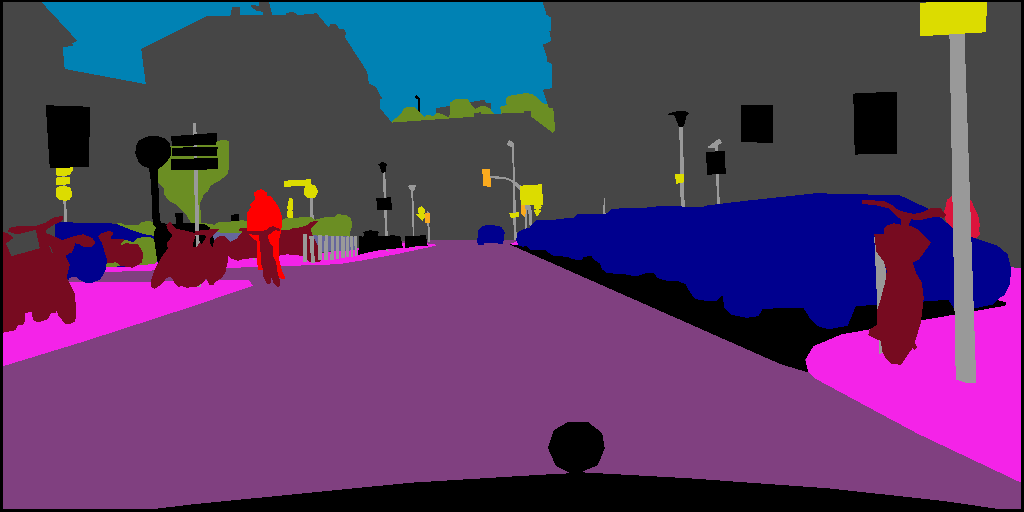}
\caption*{GT Labels}
\end{subfigure}%
\begin{subfigure}{\imgWidth}
\centering
\includegraphics[width=\textwidth]{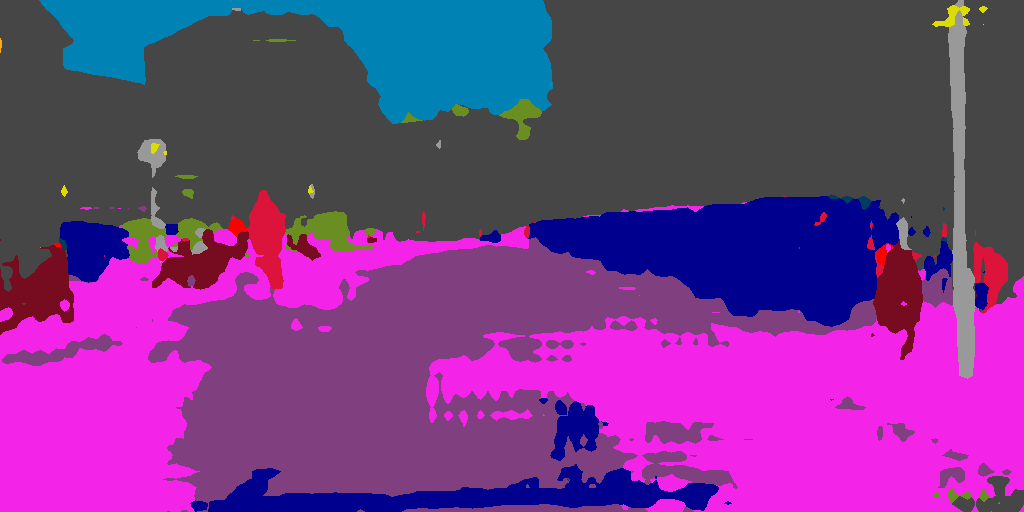}
\caption*{Baseline}
\end{subfigure}%
\begin{subfigure}{\imgWidth}
\centering
\includegraphics[width=\textwidth]{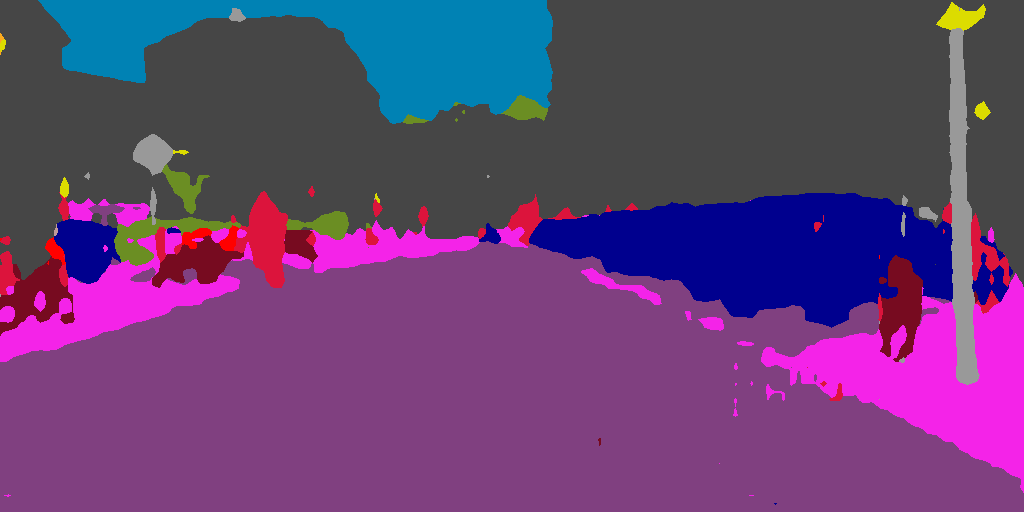}
\caption*{MaxSquareIW~\cite{Chen2019}}
\end{subfigure}%
\begin{subfigure}{\imgWidth}
\centering
\includegraphics[width=\textwidth]{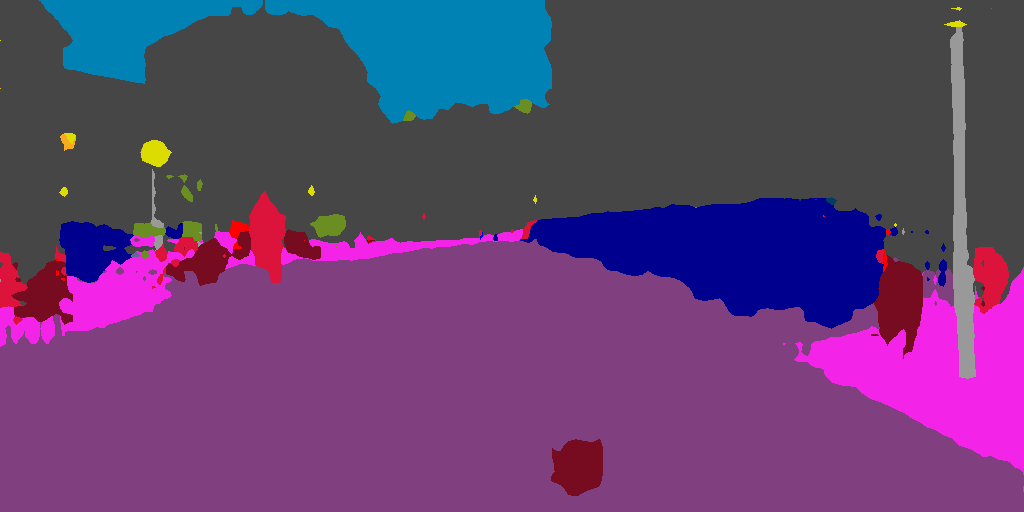}
\caption*{UDA OCE~\cite{toldo2020clustering}}
\end{subfigure}%
\begin{subfigure}{\imgWidth}
\centering
\includegraphics[width=\textwidth]{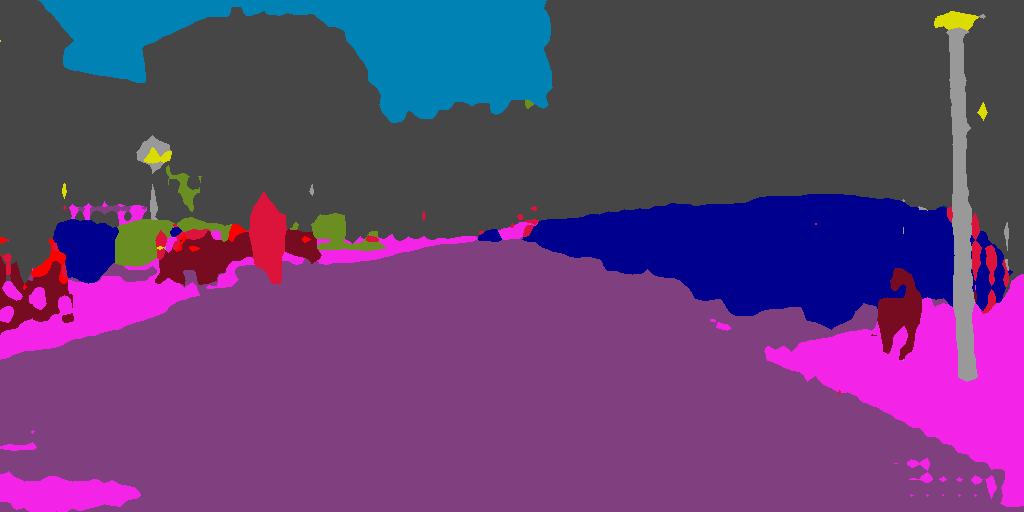}
\caption*{LSR (Ours)}
\end{subfigure}%
\end{subfigure}
\caption{Qualitative results on sample scenes taken from the \textit{Cityscapes} validation split.}
\label{fig:results}
\end{figure*}
\subsection{Adaptation from GTAV to Cityscapes}
\label{subsec:gta}
In the \textit{GTAV} $\rightarrow$ \textit{Cityscapes} setup our approach 
achieves a score of $46.0\%$ mIoU, 
with a gain of $9.1\%$ compared to the baseline.
In addition, LSR outperforms all competitors, with only the very recent works of \cite{toldo2020clustering} and \cite{Chen2019}  able to get close to our result, while there is a quite relevant gap compared to other methods.
Such improvement is quite stable across most single-class IoU scores, and is particularly evident in difficult classes, such as \textit{terrain}, 
where our strategy shows the highest percentage gains.
As an index of robustness and performance balance, 
we use the standard deviation of the per-class IoUs. LSR 
reduces it by $0.8$ 
compared to the latest state-of-the-art~\cite{toldo2020clustering} (from $24.8$ to $24.0$). 
Furthermore, LSR surpasses the 
same strategy by $0.4\%$ in terms of mASR, reaching $67.7\%$, meaning that our approach shows improved accuracy over more challenging classes, thanks to the enhanced latent space regularization. 

Some qualitative results are reported in the top half of Fig.~\ref{fig:results} and on the paper webpage \url{http://lttm.dei.unipd.it/paper\_data/LSR}. 
From visual inspection, 
we can verify the increased precision of \textit{t.\ sign}, \textit{t.\ light}, \textit{pole} and \textit{person} borders in both images. Furthermore, our approach correctly classifies the \textit{wall} in the first image (labeled as \textit{building} by competitors) and the \textit{building} in the second image (confused for \textit{fence} by competitors).
\subsection{Adaptation from SYNTHIA to Cityscapes}
\label{subsec:synthia}

In the \textit{SYNTHIA} $\rightarrow$ \textit{Cityscapes} setup LSR surpasses all the competitors in the 16-classes configuration, reaching $41.7\%$ of mIoU, with a gap compared to the best competing approach larger than in the previous setting. 
Once more, our method reduces the standard deviation of the IoU distribution ($27.7$ compared to $29.7$ of \cite{toldo2020clustering}).
At the same time, LSR shows the highest mASR, surpassing the second best approach (on 16 classes) by $2.2\%$. On the 13-classes setup our strategy outperforms all competitors in terms of mASR with only a marginal loss in mIoU score,
confirming our previous claim of better performance balance across classes and reduced gap with respect to supervised learning.

Similarly to the \textit{GTAV} case, the performance gain is visible also from the qualitative results in the bottom half of Fig.~\ref{fig:results}. 
The segmentation maps show an overall improvement in the shape of \textit{sidewalk} and \textit{pole}. 
Our method can correctly detect the \textit{car} and \textit{person} behind the \textit{pole} in the last image, which are missed or wrongly classified by competing strategies (\eg, the \textit{car} is confused as \textit{person} in \cite{Chen2019}), and can accurately predict the \textit{t.\ sign}, missed by some strategies.
Furthermore, we remark that LSR can correct the region around the car logo (bottom-part of each figure), which is often confused with \textit{bicycle}, \textit{car} or \textit{bus} by competitors.  
\begin{figure}
\begin{subfigure}{.22\textwidth}
\centering
\setlength{\tabcolsep}{1.4pt}
\linespread{1.3}\selectfont\centering
\begin{tabular}{cccc|c}
$\mathcal{L}_C$ & $\mathcal{L}_P$ & $\mathcal{L}_N$ & $\mathcal{L}_{EM}$ & mIoU \\
\toprule
& & & & 42.8 \\
& \checkmark & \checkmark & \checkmark & 44.8 \\
\checkmark & & \checkmark & \checkmark & 44.9 \\
\checkmark & \checkmark & & \checkmark & 45.2 \\
\checkmark & \checkmark & \checkmark & & 44.2 \\
\checkmark & \checkmark & \checkmark & \checkmark & 46.0\\
\end{tabular}
\vspace{-0.15cm}
\caption{}
\label{table:ablation}
\end{subfigure}%
\begin{subfigure}{.27\textwidth}
\hspace{-1.8mm}
\centering
\begin{subfigure}{0.85\textwidth}
\includegraphics[width=\textwidth]{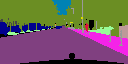}
\end{subfigure}
\begin{subfigure}{0.85\textwidth}
\includegraphics[width=\textwidth]{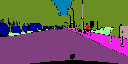}
\end{subfigure}
\vspace{-0.15cm}
\caption{}
\label{fig:downsampling}
\end{subfigure}
\vspace{-0cm}
\caption{(a) Ablation analyses. (b) Image of Fig.~\ref{fig:architecture} downsampled nearest (top) or frequency-aware (bottom).}
\end{figure}
\subsection{Ablation Studies}
\label{sec:ablation}
In this section, we 
evaluate the impact of each constraint on the final accuracy. 
Quantitative results are reported in Fig.~\ref{table:ablation}, where we evaluate our strategy by removing each constraint independently and evaluating the impact on the final accuracy. In particular, we show how the absence of each of our losses reduces the final performance by a minimum of $0.8\%$ mIoU and an average of $1\%$ mIoU. Each module brings a significant improvement in terms of accuracy and all the components are needed for the best results.
\begin{figure}
\centering
\includegraphics[trim=0.4cm 0.4cm 0.2cm 0.4cm,  clip, width=.47\textwidth]{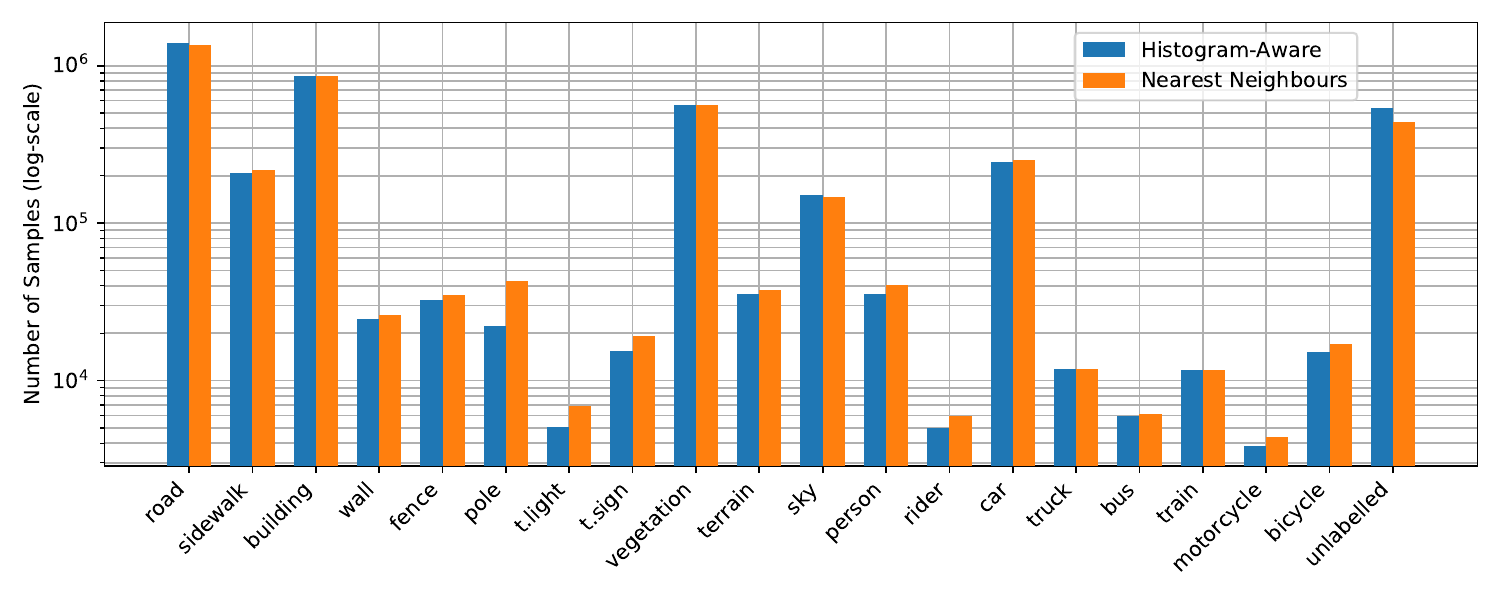}
\caption{Class distribution of segmentation maps downsampled either via histogram-aware or via nearest neighbor.}
\label{fig:hist}
\end{figure}
Concerning the novel downsampling scheme (Sec.~\ref{subsec:setup}), the goal of the proposed frequency-aware setup is to 
label only feature locations with a clear class assignment.
This aims to reduce cross-talk between neighboring features of different classes, thus improving class discriminativeness at the latent space. We can observe this phenomenon in Fig.~\ref{fig:downsampling}, where the label map downsampled via our frequency-aware scheme (bottom) marks some features close to the edges of objects as \textit{unlabeled}. This is confirmed by the class distribution of the downsampled segmentation maps (\ie, to match the spatial resolution of the feature level), reported in Fig.~\ref{fig:hist} for both the histogram-aware scheme (ours) or the standard nearest neighbors one. In particular, the histogram-aware scheme generally seldom preserves small classes, promoting \textit{unlabeled} classification when discrimination between classes is uncertain. 
\begin{figure}
\centering
\begin{subfigure}{.18\textwidth}
\centering
\includegraphics[trim=0.5cm 0.5cm 0.5cm 0.5cm, clip, width=\textwidth]{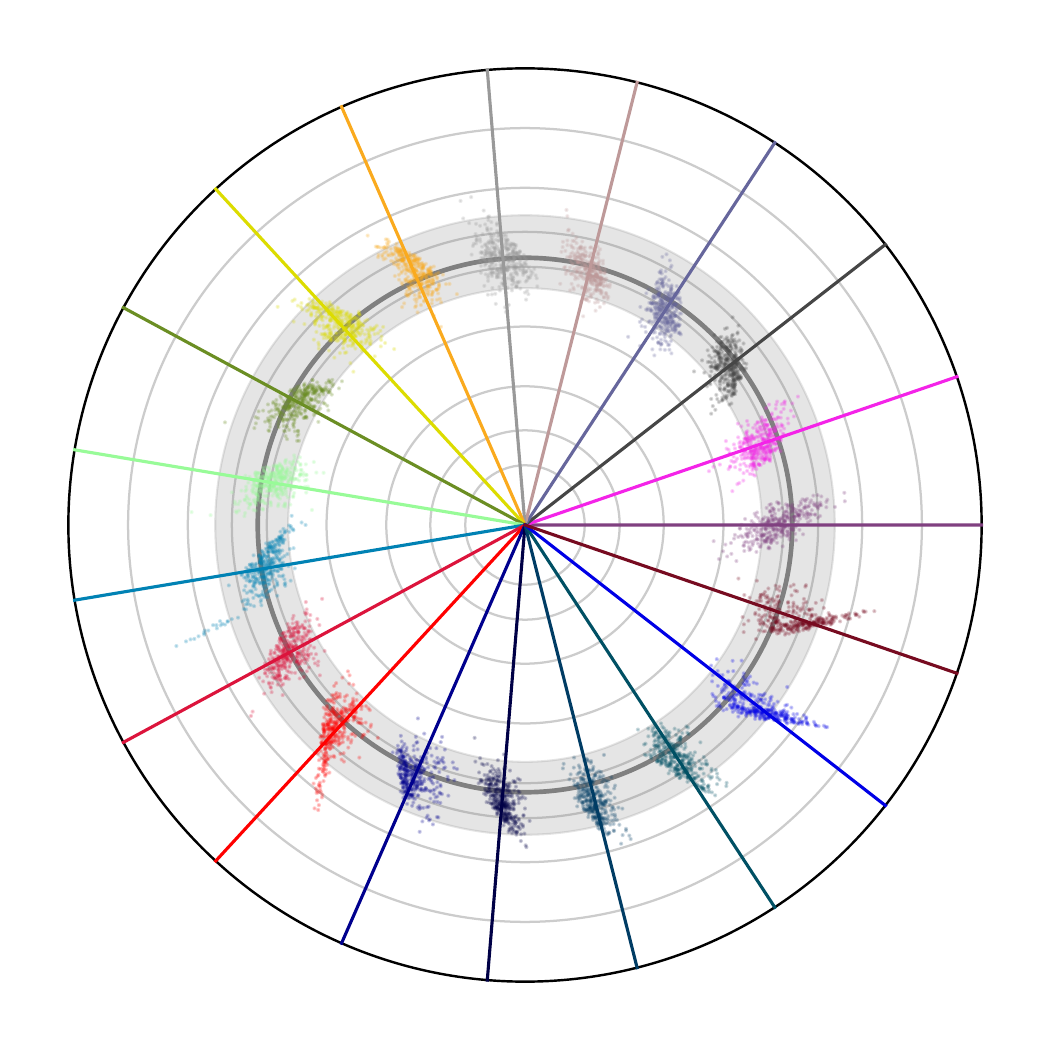}
\caption{Baseline.}
\end{subfigure}%
\begin{subfigure}{.18\textwidth}
\centering
\includegraphics[trim=0.5cm 0.5cm 0.5cm 0.5cm, clip, width=\textwidth]{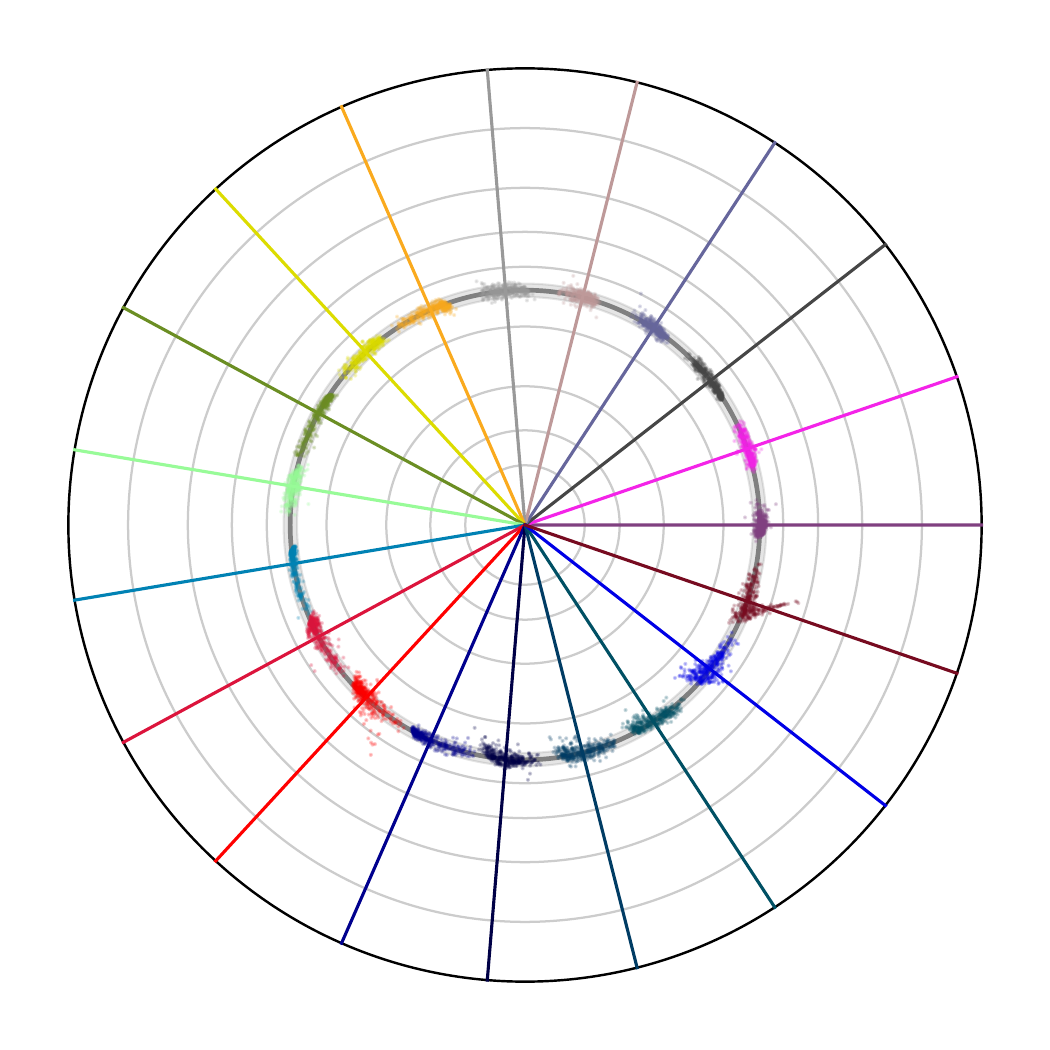}
\caption{LSR.}
\end{subfigure}
\caption{Features distribution before and after adaptation.}
\label{fig:stars}
\end{figure}


\subsection{Analysis of the Latent Space Regularization}

\textbf{Norm Alignment.} 
We analyze the effect of the norm alignment constraint in Fig.~\ref{fig:stars}, where we show a plot of some feature vectors after projecting them to a 2D space for better visualization. This and the subsequent plots were produced using a balanced subset of feature vectors ($350$ vectors per class)  extracted from the \textit{Cityscapes} validation set for a fair comparative analysis across the classes. 

\begin{figure}
\centering
\includegraphics[trim=0cm 0.5cm 0cm 0.3cm, clip, width=.47\textwidth]{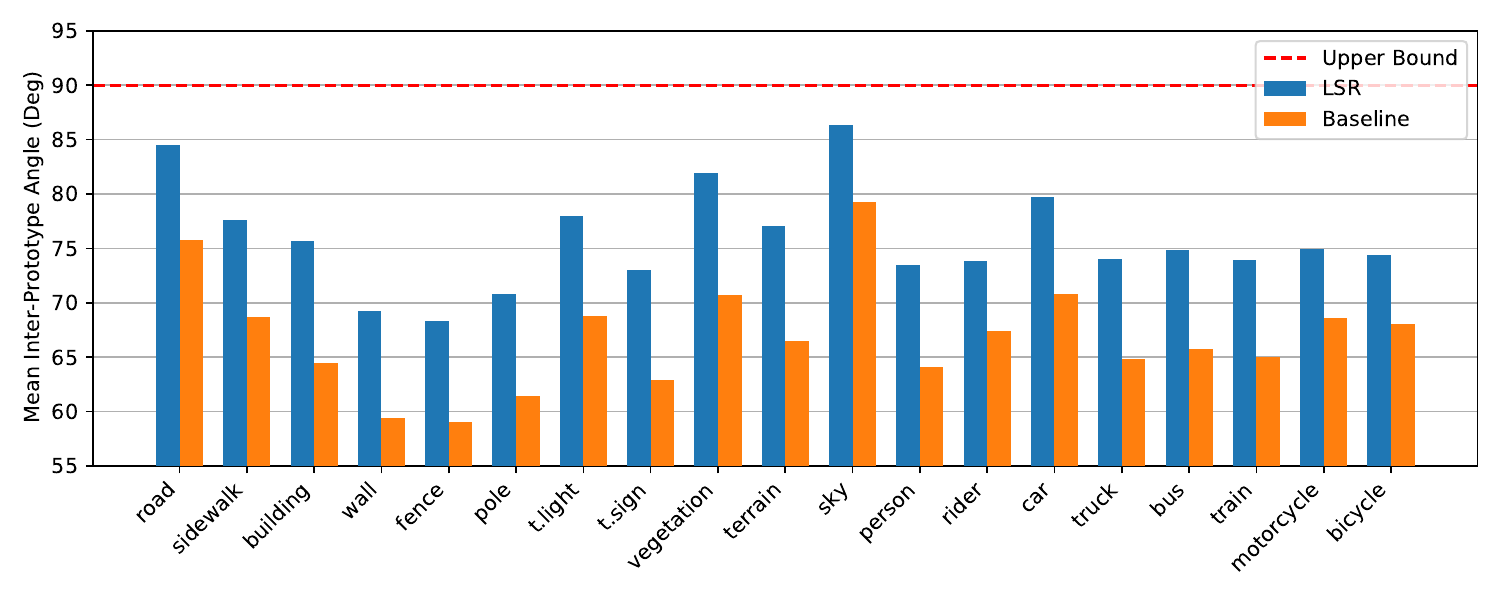}
\caption{Average inter-prototype angle, comparison between Source Only and LSR.}
\label{fig:perp}
\end{figure}

In Fig.~\ref{fig:stars} the norm of each vector is represented in the radial axis in log scale, the inner angle between any point and its prototype's direction is represented in the angular axis. 
The original direction in the high dimensional space of each centroid is ignored, as a meaningful representation would be very difficult to achieve in 2D. 
Instead, we assign to each centroid a reference angle (as shown by the colored lines) and plot 
the associated feature vectors centered on it.
The plots also reports a confidence interval for the global average norm, 
to highlight how the proposed norm alignment constraint (Sec.~\ref{loss:norm}) effectively promotes uniform norm values:
the dark gray line represents the median of the distribution and the shaded gray area represents the $95\%$ confidence interval. 
The effect of our space shaping strategy is clearly visible, in that the norms align very tightly around the global mean value (smaller shaded region around the unique gray line). The distribution of the points around each prototype is shrunk, thanks to the clustering objective, while the centroids themselves are pushed away from each other, thanks to the perpendicularity constraint (not visible from this plot, but appreciable in Fig.~\ref{fig:perp} as we discuss next).

\textbf{Perpendicularity.} 
To analyze the effect of the perpendicularity constraint, Fig.~\ref{fig:perp} shows the distribution of the average inner angle between a prototype's direction and the direction of each other prototype. Ideally, we aim at producing as perpendicular prototypes as possible, in order to reduce the overlap of different semantic classes over feature channels (\ie, cross-talk).
The red dashed line at $90$ degrees shows the target value for perpendicularity, which is also the upper bound, as our feature vectors have all non-negative coordinates.
The figure  shows that our strategy leads to an average increase of more than $5$ degrees.

\begin{figure}
\centering
\begin{subfigure}{.2\textwidth}
\centering
\includegraphics[trim=0cm 0.5cm 0cm 0.3cm, clip, width=\textwidth]{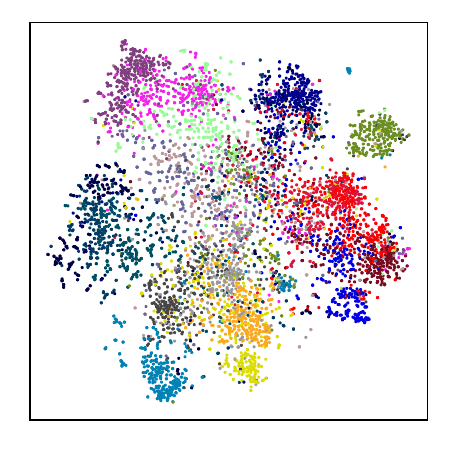}
\caption{Baseline.}
\end{subfigure}%
\begin{subfigure}{.2\textwidth}
\centering
\includegraphics[trim=0cm 0.5cm 0cm 0.3cm, clip, width=\textwidth]{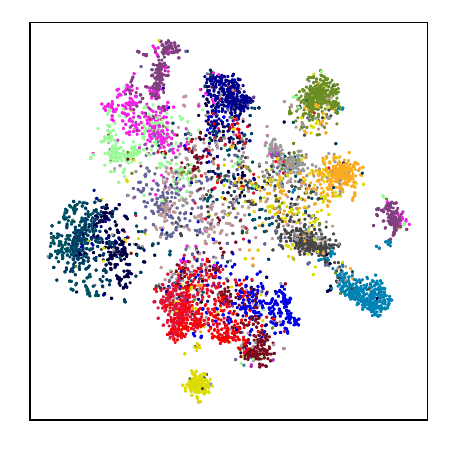}
\caption{LSR.}
\end{subfigure}
\caption{t-SNE plots comparing feature vectors distributions before and after adaptation. Points are color coded according to the legend of Fig.~\ref{fig:results}.}
\label{fig:tsne}
\end{figure}

\textbf{Clustering.} Finally, we analyze our clustering objective by means of a t-SNE~\cite{maaten2008visualizing} embedding produced on the normalized features (to remove the norm information, enhancing the angular one) and we report it in Fig.~\ref{fig:tsne}. Our strategy increases significantly the cluster separation in the high dimensional space and the spacing between clusters belonging to different classes. As a side effect, this also reduces the  probability of confusing visually similar classes (\eg, the \textit{truck} class with the \textit{bus} and \textit{train} ones). 

\section{Conclusions}
\label{subsec:conclusions}
In this work, we proposed a new set of latent-space regularization techniques to address the domain shift in an unsupervised fashion. We achieve domain invariance by means of multiple feature space shaping constraints: namely, class clustering, class perpendicularity and norm alignment. Our constraints can be flawlessly applied on top of existing frameworks as they are separate modules trained end-to-end.
We achieved state-of-the-art results in feature-level adaptation on two commonly used benchmarks 
paving the way to employment of a new family of feature-level techniques  to  enhance discrimination ability of deep neural networks. Future research will concern the design of novel feature-level techniques, the analysis of the  proposed adaptation strategies on source accuracy and the evaluation of their generalization ability to different tasks.

\newpage
{\small
\bibliographystyle{ieee_fullname}
\bibliography{strings,refs}
}

\end{document}